\documentclass{artifactpreprint}
\usepackage{fvreport}
\newcommand{\fv}{FrontierVideo\xspace}

\newcommand{\catfootage}{\texttt{footage}\xspace}
\newcommand{\catmotion}{\texttt{motion}\xspace}
\newcommand{\catcamera}{\texttt{camera3d}\xspace}

\graphicspath{{figures/shots/}{figures/}}
\definecolor{catV}{HTML}{0064E0}
\definecolor{catF}{HTML}{D9730D}
\definecolor{catM}{HTML}{168A6B}
\definecolor{catC}{HTML}{7446B8}
\newlength{\fvshotw}\newlength{\fvshoth}
\newcommand{\fvshotratio}{0.5625}
\newif\iffvthirdparty \fvthirdpartytrue
\newcommand{\shot}[3][]{%
  \def\fvtparg{#1}%
  \ifx\fvtparg\empty\includegraphics[width=#2]{#3}\else
    \iffvthirdparty\includegraphics[width=#2]{#3}\else
      \setlength{\fvshotw}{#2}\setlength{\fvshoth}{\fvshotratio\fvshotw}%
      \fcolorbox{metalight}{metabg}{\parbox[c][\dimexpr\fvshoth-2\fboxsep-2\fboxrule\relax][c]{\dimexpr\fvshotw-2\fboxsep-2\fboxrule\relax}{\centering\tiny\color{metagray}third-party frame withheld}}%
    \fi\fi}

\newcommand{\slug}[1]{{\fontsize{6}{7}\selectfont\ttfamily\color{metagray}#1}}

\title{\fontsize{20}{24}\selectfont\color{fvBrandViolet}\fv: Benchmarking Coding Agents\\for Professional Video Production}

\author{%
  \fontsize{14}{17}\selectfont
  \textcolor{black!85}{FrontierVideo Bench Team}}

\preprintbrand{\includegraphics[width=\linewidth]{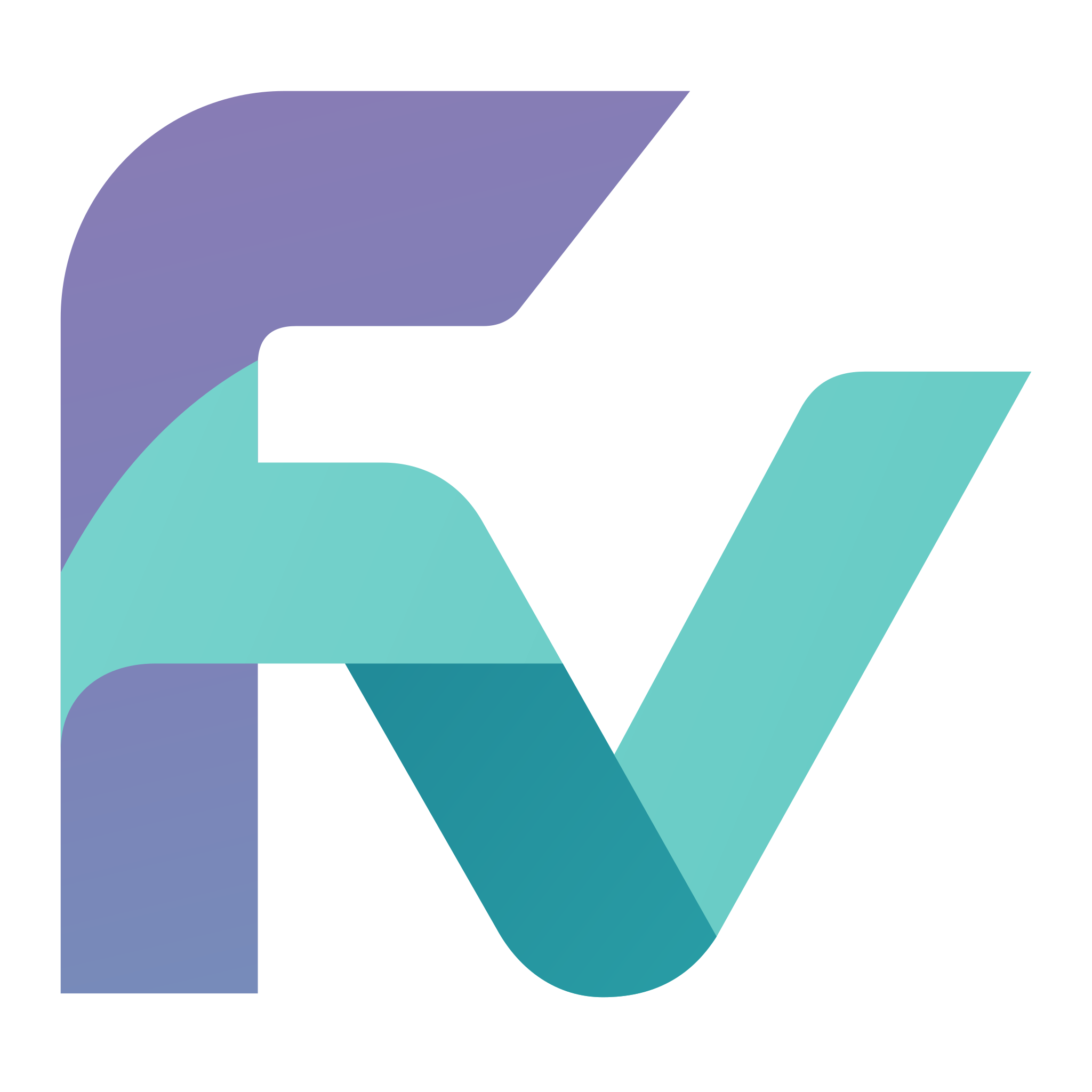}}
\preprintmeta{}
\preprintlinks{%
  \begin{tabular}{@{}l@{\hspace{1.1em}}l@{}}
    \href{https://frontiervideo.github.io/}{%
      \raisebox{-0.10em}{\faGlobe}\hspace{0.45em}\textbf{Project Website}} &
    \href{https://frontiervideo.github.io/}{frontiervideo.github.io}\\[0.35em]
    \href{https://github.com/FrontierVideo/FrontierVideo}{%
      \raisebox{-0.10em}{\textcolor{black}{\faGithub}}\hspace{0.45em}\textbf{GitHub Repo}} &
    \href{https://github.com/FrontierVideo/FrontierVideo}{github.com/FrontierVideo/FrontierVideo}\\[0.35em]
    \href{https://huggingface.co/datasets/FrontierVideo/frontiervideo-media-public}{%
      \mbox{\raisebox{-0.20em}{\includegraphics[height=1.1em]{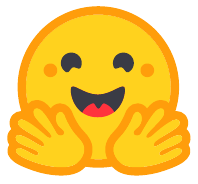}}%
        \hspace{0.45em}\textbf{HuggingFace~Dataset}}} &
    \href{https://huggingface.co/datasets/FrontierVideo/frontiervideo-media-public}{huggingface.co/datasets/FrontierVideo/frontiervideo-media-public}\\[0.35em]
    \href{mailto:tianfuwang.cs@gmail.com}{%
      \raisebox{-0.10em}{\faEnvelope}\hspace{0.45em}\textbf{Contact Emails}} &
    \href{mailto:tianfuwang.cs@gmail.com}{tianfuwang.cs@gmail.com}%
    \quad\textcolor{preprintMuted}{\textperiodcentered}\quad
    \href{mailto:xionghui@ust.hk}{xionghui@ust.hk}
  \end{tabular}%
}

\preprintabstract{%
Coding agents are extending digital artifact creation from documents and websites to video.
Professional video production requires precise control over appearance, layout, timing and motion, yet existing benchmarks struggle to verify these fine-grained requirements reliably.
We introduce \textbf{\fv}, a benchmark of 100 tasks across five production categories, with 20 in each. Each task requires an editable source project and the video it renders, and \fv grounds its evaluation in the objects that the task specifies.
The \emph{Reconstruction} track covers Visual Recreation, Footage Editing, Motion Fusion and 3D Camerawork. \fv executes each submitted project and uses a public object contract to verify object-level properties against a reference.
The \emph{Creation} track comprises Script Realization tasks, in which a vision-language model (VLM) judge answers screenplay-derived questions about the objects in each shot while design choices remain open.
A five-stage pipeline authors each task, fixes its scoring plan before any agent is evaluated and validates it with controls.
We evaluate seven agent systems powered by frontier models, each given six hours per task.
The strongest system reaches 60.9/100 overall, and none exceeds 26.4 on Motion Fusion. Object-level scores show where systems lose credit, from missing targets in Footage Editing and Motion Fusion to camera framing in 3D Camerawork, and the two tracks rank systems differently.}

\hypersetup{%
  pdftitle={FrontierVideo: Benchmarking Coding Agents for Professional Video Production},
  pdfauthor={FrontierVideo Bench Team},
  pdfsubject={Coding agents for video engineering: Reconstruction and Creation benchmark design}%
}

\begin{document}
{\colorlet{preprintPanel}{fvBrandViolet!14!white}%
\maketitle}

\vspace*{-6pt}\noindent\begin{minipage}{\linewidth}
\centering
\includegraphics[width=.92\linewidth]{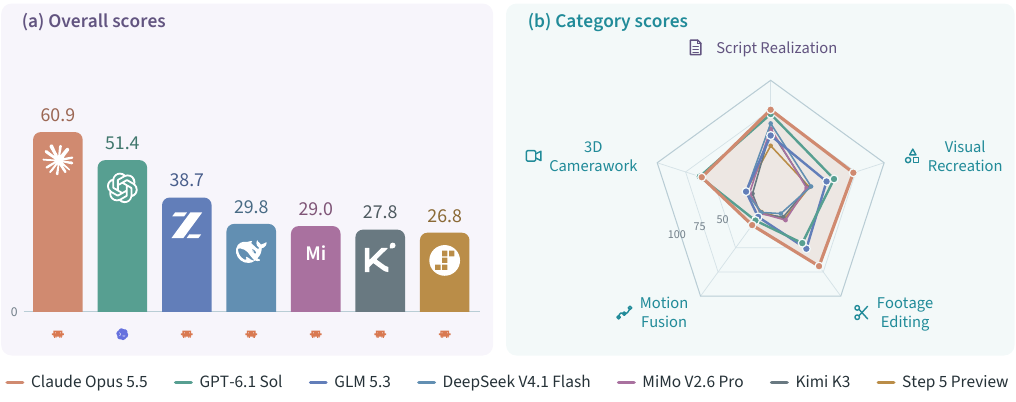}
\captionsetup{skip=4pt}
\captionof{figure}{\textbf{Overall and category scores.} (a) Overall score across the five categories for seven agent systems, with 20 tasks per category. (b) Category scores, in the colors of (a); the three systems with the highest overall scores are drawn with thicker lines. Reconstruction uses offline regrades of the archived submissions under their recorded scoring plans. Script Realization combines the \href{https://frontiervideo.github.io/?leaderboard=create\&lbform=script\#leaderboard}{original 20-task study} with later runs on the same tasks. \Cref{tab:controlled-results} lists the scores.}
\label{fig:leaderboard-layout-preview}
\end{minipage}
\clearpage

\noindent\begin{minipage}{\linewidth}
\centering
\iffvthirdparty
  \includegraphics[width=\linewidth]{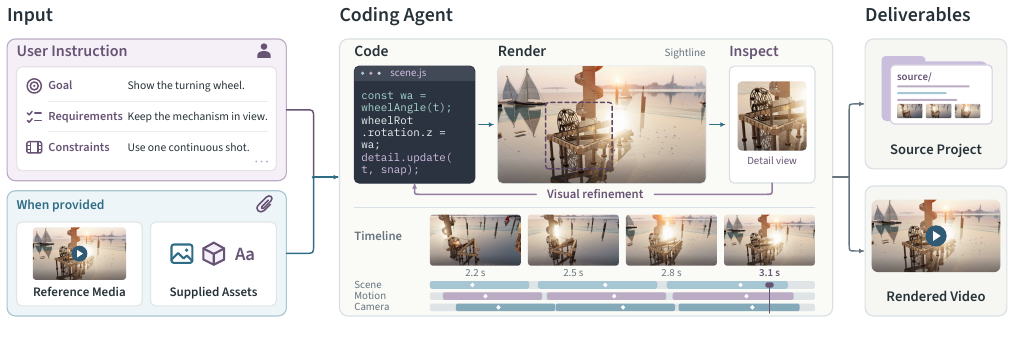}
\else
  \includegraphics[width=\linewidth]{fig-benchmark-overview-public.pdf}
\fi
\captionsetup{skip=6pt}
\captionof{figure}{\textbf{Code-to-video production workflow.} The user instruction specifies a goal, requirements and constraints. \mbox{Reference media} and production assets are supplied as appropriate. A coding agent iterates through code, rendering and inspection to deliver an editable source project and rendered video. The schematic workspace uses a code excerpt, timestamped frames and a matching detail crop from the \emph{Sightline} Motion Fusion reference.}
\label{fig:two-tracks}
\end{minipage}\par

\section{Introduction}
\label{sec:intro}

Coding agents increasingly produce complete digital artifacts, from documents and websites to visual media~\citep{wang2026agentic}. Code-based video systems already write a scene, render it and revise the source after inspecting the result~\citep{chen2025code2video}, as illustrated in \cref{fig:two-tracks}. Professional production raises the bar. In a product reveal, the camera may need to show a component only after its cover slides away, while a label stays visible until the cut. Such requirements bind individual elements to precise positions, times and motions. A film can therefore look right overall yet fail the brief, for example when the cover opens a second late or the label disappears before the cut.

Evaluating these requirements is the central difficulty. Video benchmarks for coding agents mostly judge the rendered output, through frame metrics, vision-language model (VLM) judges or learned preference models~\citep{zala2026code2videobenchmark,chen2026code2videocompass,gupta2026timelinebench}. Frame metrics favor per-frame appearance over motion~\citep{ge2024fvdcontentbias}, and a single similarity score cannot show which element failed or how, leaving the failure patterns of agents hidden. VLM judges are bounded by the perception of current models, which remains unreliable for temporal order and precise motion such as camera trajectories~\citep{li2026isovrbench,lin2025camerabench}. This raises the question of how element-level production requirements can be verified reliably across different code implementations.

\Needspace{5\baselineskip}
We introduce \fv, an object-grounded code-to-video benchmark that verifies the production requirements of a task specification. As \cref{tab:production-capabilities} summarizes, it comprises five production categories of 20 tasks, each targeting a core capability of video production: building what appears in the frame (Visual Recreation), arranging supplied media (Footage Editing), coordinating how elements move together (Motion Fusion), filming a 3D scene (3D Camerawork) and turning a screenplay into a complete film (Script Realization). The categories fall into two tracks according to how the goal is given. In \emph{Reconstruction}, the first four categories give it as a reference video whose specified properties the submission must reproduce. In \emph{Creation}, Script Realization gives it as a screenplay, which fixes the required content, actions and timing but leaves unprescribed design choices open. In every task, the agent delivers an editable source project and the video it renders. The primary outcome, \emph{Faithful Realization}, measures how much of the specified work a submission delivers. It reflects both how well the agent understands the task inputs, perceiving the reference video or separating a screenplay's obligations from its open choices, and how well it implements them in code; a single score does not separate the two. \emph{Expressive Quality}, how well the film's choices serve its purpose, is a secondary outcome.

For Reconstruction, \fv evaluates a video element by element. Each video is decomposed into objects, such as graphic elements, material instances, motion units and cameras. The task inputs give each object an ID and, where applicable, a required code name or a boundary mask. The agent binds its content to those IDs. Reference values, tolerances and weights stay private. The evaluator then reads object properties from the executed reference and submitted projects and compares them under the same scoring plan. This lets different implementations earn the same credit when they satisfy the same checks. Creation leaves the design open, so a VLM judge answers binary screenplay-derived questions on clips of the rendered film.

Tasks are built through a five-stage authoring pipeline that yields a feasible reference project, a public task package and private scoring checks. Validation then tests the checks themselves: blank, frozen and fault-injected outputs probe whether they detect missing content, missing motion and attribute errors, and valid alternatives test whether they respect the choices a brief leaves open. Four tasks per category are public, and the remaining 80 are held out.

\Needspace{5\baselineskip}
We evaluate seven complete agent systems, each with a fixed model, harness, tools and feedback loop. Claude Opus~5.5 leads with an Overall score of 60.9, followed by GPT-6.1 Sol at 51.4. Motion Fusion yields the lowest category mean for every system. Decomposing scores by object and property shows where points are lost: many targets earn no credit in Footage Editing and Motion Fusion, and five systems score much lower on spatial checks than on motion checks in 3D Camerawork.

\Needspace{11\baselineskip}
Our contributions are summarized as follows:
\begin{itemize}[leftmargin=1.4em,topsep=0.2em,itemsep=0.2em,parsep=0pt,partopsep=0pt]
\item \textbf{Benchmark design.} We introduce 100 tasks across five video production capabilities. Each task gives a reference video (Reconstruction) or a screenplay (Creation) and requires an editable project and its render.
\item \textbf{Object-grounded evaluation.} Public object IDs align targets across implementations. We execute the reference and submitted projects and compare their observed properties under a fixed scoring plan.
\item \textbf{Authoring pipeline.} Five stages link production briefs, reference projects, task packages and grading evidence. Validation tests whether the checks detect faults and preserve credit for valid alternatives.
\item \textbf{Empirical findings.} The best of seven systems reaches 60.9/100 overall, and none exceeds 26.4 on Motion Fusion. Object-level scores show how lost points vary by target and property group.
\end{itemize}

\section{Related Work}
\label{sec:related}

\afterpage{%
\noindent\begin{minipage}{\linewidth}
\centering\small\color{black}
\setlength{\tabcolsep}{4.5pt}
\renewcommand{\arraystretch}{1.14}
\captionof{table}{\textbf{Benchmark comparison.} Goal inputs, production scope and principal evaluation mechanisms of closely related video benchmarks. Reference world-state metrics in 4DCodeBench apply to synthetic scenes.}
\label{tab:benchmark-comparison}
\resizebox{\linewidth}{!}{%
\fontsize{10}{12}\selectfont
\begin{tabular}{@{}ll>{\raggedright\arraybackslash}p{5.6cm}>{\raggedright\arraybackslash}p{7.8cm}@{}}
\toprule
\tabhead{Benchmark} & \tabhead{Goal input} & \tabhead{Production scope} & \tabhead{Evaluation basis} \\
\midrule
Code2Video (HeyGen)~\citep{zala2026code2videobenchmark} & Textual brief & Code-rendered motion design & Pairwise video judge trained on human preferences across five design axes \\[5pt]
Code2Video-Compass~\citep{chen2026code2videocompass} & Text, image or video & General code-rendered video & Execution, video quality, alignment and reference metrics, plus model-judged checklists \\[5pt]
Timeline-Bench~\citep{gupta2026timelinebench} & Client brief and footage & Editing supplied production material & Delivery and content tests, brief compliance and an editor-calibrated video-quality bar \\[5pt]
4DCodeBench~\citep{shen2026fourDcodebench} & Reference video & Dynamic 3D scene reconstruction & Reference-video and world-state metrics, plus visual question answering and pairwise judgments \\[5pt]
PhysCodeGym~\citep{shao2026physcodegym} & Reference video & Physics-based Blender reconstruction & Visual, event and motion scores, program and scene inspection, plus selected counterfactual tests \\
\midrule
\rowcolor{fvBrandCyan!8}
\textcolor{fvBrandTeal}{\textbf{FV-Reconstruction}}
& Reference video & 80 tasks: frame, media, motion, camera & Object-level observations from executed projects, matched through a public object contract \\[3pt]
\rowcolor{fvBrandViolet!7}
\textcolor{fvBrandViolet}{\textbf{FV-Creation}}
& Textual screenplay & 20 tasks: Script Realization & Binary judgments of the rendered film against screenplay-derived criteria \\
\bottomrule
\end{tabular}%
}
\end{minipage}\par

\par\vspace{10pt}
\afterpage{\noindent\begin{minipage}{\linewidth}
\centering
\iffvthirdparty
  \includegraphics[width=\linewidth]{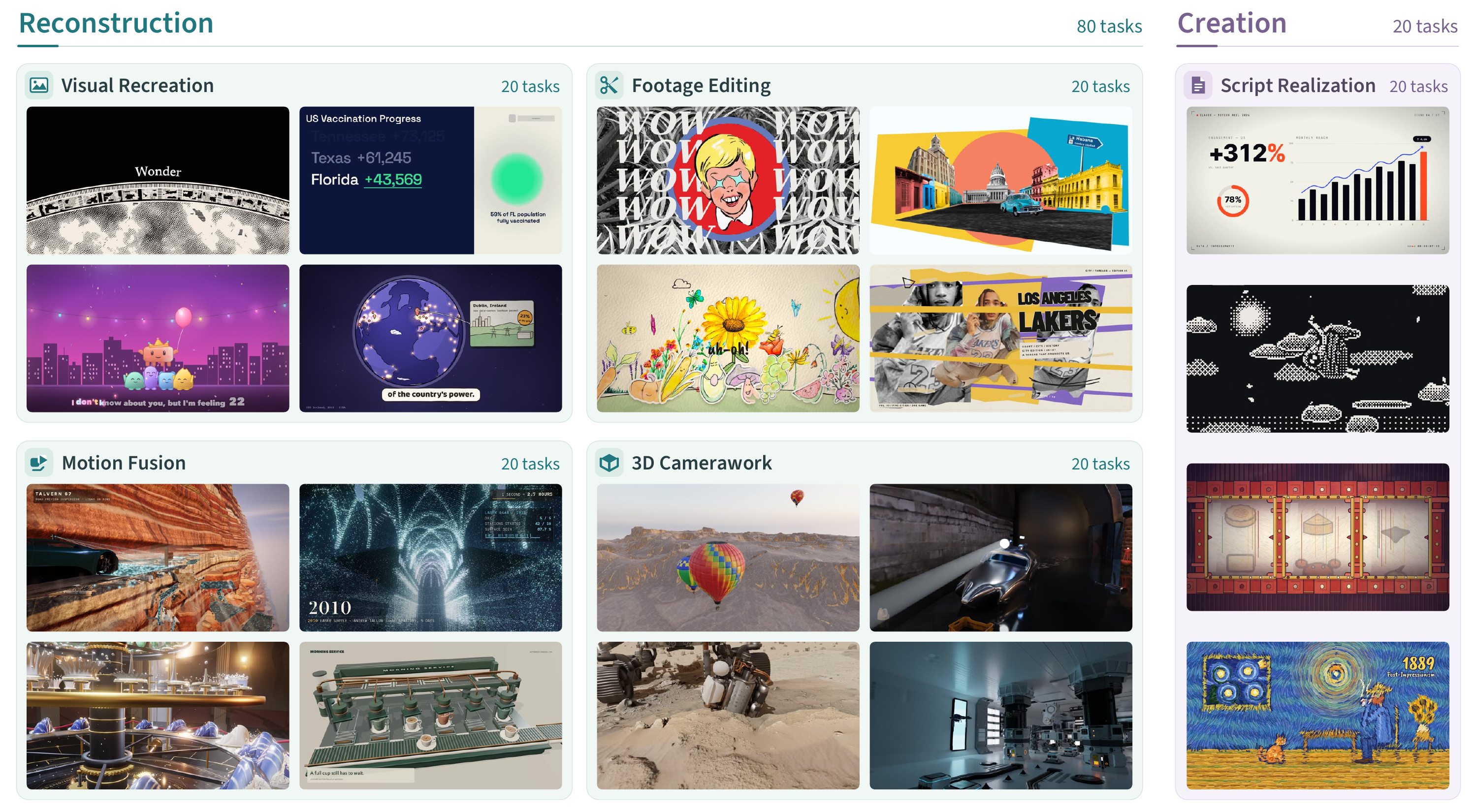}
\else
  \includegraphics[width=\linewidth]{figures/fig-task-gallery-public.pdf}
\fi
\captionsetup{skip=6pt}
\captionof{figure}{\textbf{Five production categories in FrontierVideo.} Each category contains 20 tasks and shows four examples. Reconstruction comprises four categories (80 tasks), while Creation comprises Script Realization (20 tasks). From top to bottom, the Creation panels show brief-development sources for \emph{CLAUDE~\textperiodcentered{} Graphic Reel}, \emph{Scarf Bird}, \emph{Moon Missing a Corner}, and \emph{Cat \& Coffee}. All four are among the 20 Script Realization tasks, and agents receive only the text briefs. Construction details are in \cref{app:construction-details}.}
\label{fig:gallery}
\end{minipage}\par
}%
}

\subsection{Coding Agent Benchmarks}

Coding agent benchmarks typically verify submissions by execution in isolated environments, from repository issues~\citep{jimenez2024swebench,miserendino2025swelancer} to web and desktop control~\citep{zhou2024webarena,xie2024osworld} and machine-learning engineering~\citep{chan2025mlebench}.
Terminal-Bench~\citep{merrill2026terminalbench} makes this practice explicit by packaging each command-line task with its environment, a reference solution and tests in the \href{https://docs.harborframework.com/tasks/overview}{Harbor task format}.
As agents increasingly produce visual artifacts, evaluation has extended to static web pages, charts and vector graphics~\citep{si2024design2code,yang2025chartmimic,rodriguez2023starvector}, to screenshots in software-engineering tasks~\citep{yang2025swebenchmm} and to engine tests of game behavior~\citep{chi2026gamedevbench}.
For artifacts that change over time, however, the evidence usually remains visual. ArtifactsBench checks screenshots sampled at specified times against checklists~\citep{zhang2026artifactsbench}, while GDPval~\citep{patwardhan2026gdpval} and the Remote Labor Index~\citep{mazeika2025remotelabor}, which include film, video and animation work, grade whole deliverables by blinded expert comparison and by whether a reasonable client would accept them, respectively.
Execution-based checks are also fallible: audits report tests that reject correct solutions, missing checks and training-data contamination~\citep{chowdhury2024swebenchverified,openai2026swebenchverified,zhu2025abc}, and frontier models exploit tests altered to contradict the specification~\citep{zhong2026impossiblebench}.
\fv grades Reconstruction from executed projects and tests its checks with fault-injected outputs and valid alternatives.

\subsection{Code-Based Video Generation}

Video generation models such as Sora, Veo~3 and Movie Gen synthesize footage directly as pixels~\citep{brooks2024sora,deepmind2025veo3,polyak2024moviegen}.
Code-based systems instead write programs that render the video, which keeps the result editable. TheoremExplainAgent~\citep{ku2025theoremexplainagent} and the Manim-based Code2Video system~\citep{chen2025code2video} produce explainers, LogoMotion~\citep{liu2025logomotion} animates layered logos, and MoVer~\citep{ma2025mover} checks SVG motion with logic predicates.
Beyond two-dimensional animation, agents script 3D scenes in Blender~\citep{hu2024scenecraft,gu2025blendergym}, plan camera trajectories and cutscenes~\citep{zhang2026filmflow,he2026cutsceneagent}, rebuild real videos as animated Blender scenes~\citep{tang2026bvb} and edit supplied footage~\citep{sandovalcastaneda2025editduet,gupta2026timelinebench,cao2026agenticvbench}.
Benchmarks for this setting evaluate code-generated video conditioned on text, images or video~\citep{chen2026code2videocompass} and brief-driven motion design~\citep{zala2026code2videobenchmark}, as well as the reconstruction of web animations~\citep{ji2026animation2code} and physical dynamics~\citep{shen2026fourDcodebench,shao2026physcodegym} from video.
Several of them grade the rendered video: Timeline-Bench applies tests and a calibrated panel of model judges to the delivered file~\citep{gupta2026timelinebench}, Code2Video-Compass compares rendered frames and CoTracker3 point tracks with image or video references~\citep{chen2026code2videocompass}, and the HeyGen benchmark ranks videos with a learned pairwise judge~\citep{zala2026code2videobenchmark}.
Camera evaluation is especially limited. CinemaBench has no reference trajectory for its text-directed tasks~\citep{zhang2026filmflow}, 4DCodeBench has no camera metric and aligns reconstructions with the reference by the iterative closest point algorithm before computing 3D metrics~\citep{shen2026fourDcodebench}, and BVB scores renders through question answering and latent similarity~\citep{tang2026bvb}.
PhysCodeGym~\citep{shao2026physcodegym} goes further by inspecting the submitted Blender program and scene, through an agent verifier with annotated rubrics and counterfactual re-simulation on selected tasks.
\fv instead reads object properties from executed projects and aligns implementations through a public object contract.

\subsection{Video Quality Evaluation}

Generated video is commonly scored with distributional metrics such as Fr\'echet Video Distance~\citep{unterthiner2018fvd}, which can favor per-frame appearance over motion~\citep{ge2024fvdcontentbias}.
Multidimensional suites broaden the assessment to quality, prompt faithfulness and compositional binding with automatic evaluators~\citep{huang2024vbench,sun2024t2vcompbench}, yet low-motion videos can still obtain high consistency scores~\citep{liao2024devil}.
Learned and rubric-grounded judges approximate human ratings more closely~\citep{he2024videoscore,zeng2026videoargus}, but they struggle to tell generated videos from frame-reordered copies~\citep{li2026isovrbench}.
Camera motion poses a similar problem: structure-from-motion models miss motions defined by scene content, while vision-language models miss precise trajectories~\citep{lin2025camerabench}.
Large-scale systems therefore still rely on people. Movie Gen uses pairwise human evaluation because existing automated metrics gave unreliable results, and rates visual quality independently of the text prompt~\citep{polyak2024moviegen}.
For agentic artifacts more broadly, \citet{wang2026agentic} distinguish requirement satisfaction, artifact quality, effectiveness in use and internal consistency.
\fv separates fulfillment of stated production requirements from whole-film Expressive Quality.

\Cref{tab:benchmark-comparison} compares closely related benchmarks by their goal inputs, production scope and evaluation basis.

\Needspace{11\baselineskip}
\section{The FrontierVideo Benchmark}
\label{sec:overview}
\suppressfloats[t]

\fv assesses each requirement at the level at which the task specifies it. Reconstruction requirements are grounded in named objects and checked against properties read from the executed project, while Creation requirements are checked on the rendered film. We first define the tasks (\cref{sec:task-formulation}) and the protocol that scores them (\cref{sec:verification}). We then describe how authoring fixes each task's scoring plan before any agent is evaluated (\cref{sec:construction}) and summarize the resulting collection (\cref{sec:overview-scale}).

\begin{figure}[!t]
\centering
\includegraphics[width=\linewidth]{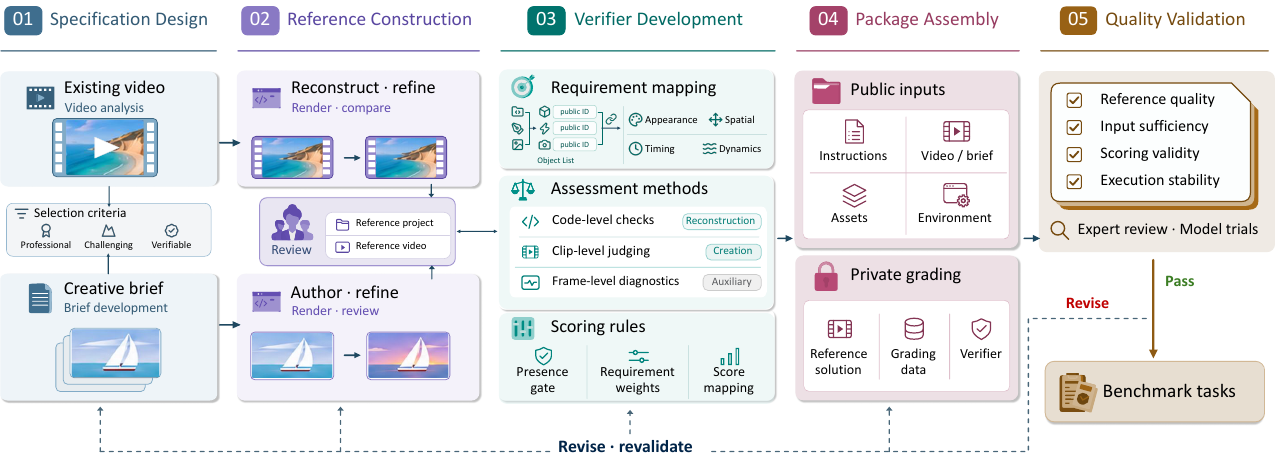}
\caption{\textbf{Stages of task authoring.}
(01)~Select a source video or draft a creative brief.
(02)~Build and refine an editable reference project.
(03)~Map requirements to checks and define scoring rules.
(04)~Package public inputs and private grading files separately.
(05)~Validate the task through expert review and model trials, then revise as needed.}
\label{fig:pipelines}
\end{figure}

\subsection{Task Definition}
\label{sec:task-formulation}
\label{sec:tracks}
Each task is a self-contained \href{https://docs.harborframework.com/tasks/overview}{Harbor package} that provides a goal specification, either a reference video or a screenplay, together with starter files, any supplied assets and a pinned execution environment.
A public manifest separates reusable assets from reference videos, frames and masks supplied for inspection only.
The agent works with code-based production tools, may not call image- or video-generation models, and submits an editable project together with its rendered video.
Grading data remain outside the agent's environment.

\paragraph{Reconstruction.}
The agent reproduces the specified visual and temporal properties of a reference video.
The four categories isolate different production demands: building the frame, arranging media, coordinating motion and filming with a camera.
Visual Recreation rebuilds nearly every element in code, even when fonts or icons are supplied. Footage Editing tests the selection, placement and timing of supplied images, vector graphics and fonts. Motion Fusion concentrates on relations among moving objects. 3D Camerawork fixes the scene so that only the camera remains to be programmed.
A public object list names the objects to be assessed, and the agent binds them to the content its project draws (\cref{sec:structured-evidence}).
Requirements on asset identity, part structure and editability are checked in the project, so two similar-looking renders can score differently.

\Needspace{5\baselineskip}
\paragraph{Creation.}
The agent realizes a textual screenplay or detailed shot brief.
Required content, actions, transitions and timing define the obligations, while permitted visual choices remain open, so several designs may satisfy the same screenplay.

\Cref{tab:production-capabilities} summarizes what each category tests and what typically makes it difficult, and \cref{tab:category-specs} lists its inputs, deliverable and reference.
Because these capabilities overlap, each task is grouped by its main production demand and records the others as tags, as the examples in \cref{fig:gallery} show.

\begin{table}[!t]
\centering\small
\renewcommand{\arraystretch}{1.15}
\caption{\textbf{Core production capabilities and typical challenges.}}
\label{tab:production-capabilities}
\setlength{\tabcolsep}{7pt}
\resizebox{\linewidth}{!}{%
\begin{tabular}{@{}llll@{}}
\toprule
\tabhead{Track} & \tabhead{Task category} & \tabhead{Capability assessed} & \tabhead{Typical source of difficulty} \\
\midrule
\textcolor{fvBrandTeal}{Reconstruction}
  & Visual Recreation & Recreate layout, type, color and hierarchy
  & Alignment, styling and layering in dense frames \\[2pt]
  & Footage Editing & Select, sequence and composite given media
  & Matching asset crops, layer order and timing \\[2pt]
  & Motion Fusion & Coordinate motion across interacting elements
  & Coupled paths, occlusion and synchronized events \\[2pt]
  & 3D Camerawork & Frame subjects and reproduce camera motion
  & Pose and field of view set framing and parallax \\
\midrule
\textcolor{fvBrandInkViolet}{Creation}
  & Script Realization & Turn a screenplay into a complete film
  & Timing and continuity while staging each shot \\
\bottomrule
\end{tabular}%
}
\end{table}\par

\paragraph{Video Purposes.}
Whereas categories describe the work the agent performs, purposes describe what the film is for. We label each film by its intended use (\emph{Promotion}, \emph{News}, \emph{Education}, \emph{How-to}, \emph{Analysis}, \emph{Storytelling}, \emph{Art} or \emph{Recap}). An educational film, for example, may require visual reconstruction, footage editing or camerawork.

\FloatBarrier
\Needspace{8\baselineskip}
\subsection{Evaluation Protocol}
\label{sec:verification}

\begin{figure}[!t]
\centering
\includegraphics[width=\linewidth]{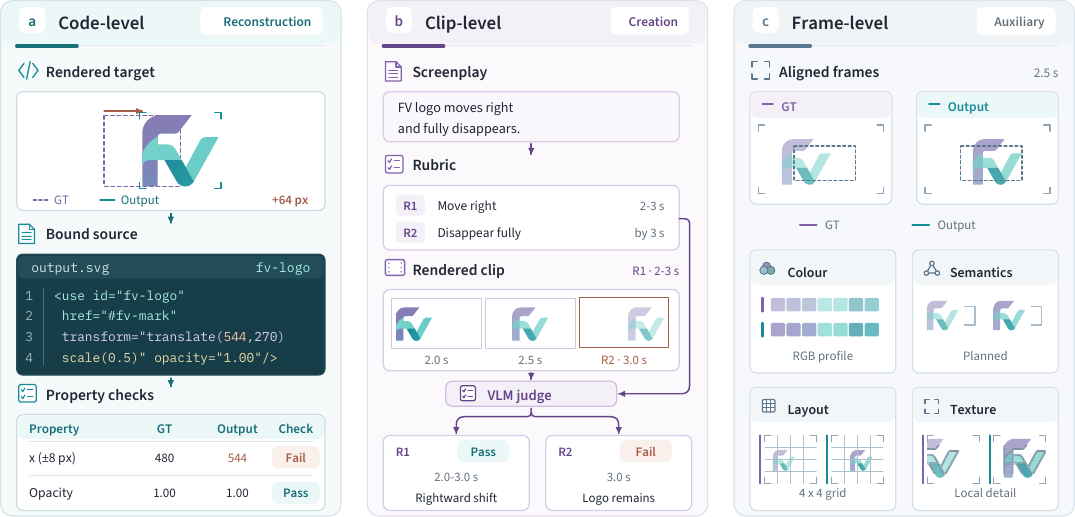}
\caption{\textbf{Three assessment mechanisms.} (a) A 64 px position offset fails an illustrative 8 px tolerance, while opacity matches. (b) A VLM judge assesses screenplay criteria using a continuous clip. Example records link rightward-motion success to 2.0--3.0 s and disappearance failure to the visible logo at 3.0 s. (c) Aligned frames support separate auxiliary diagnostics. Panels (a) and (b) supply task scores. The scripted SVG example illustrates the mechanisms, not measured agent results or actual model judgments.}
\label{fig:assessment-mechanisms}
\end{figure}

The primary score measures \emph{Faithful Realization}, the degree to which the submitted project fulfills the task's explicit production requirements. Whole-film \emph{Expressive Quality} is a secondary outcome reserved for a separate human study (\cref{sec:future-work}).
Each requirement applies to an \emph{assessment target}, which may be an object, a group of objects, an event or a relation, as listed in \cref{tab:assessment-targets}. A requirement is checked through one or more \emph{criteria}, each specifying a property, a temporal scope and a scoring rule.
As \cref{fig:assessment-mechanisms} illustrates, the two tracks differ in what counts as evidence.
Reconstruction executes the submitted project and the private reference project behind the reference video, which serves as ground truth (GT), and compares their structured observations, so a prescribed trajectory is checked against the executed camera or object states.
Creation asks a VLM judge whether each screenplay criterion is visible in the rendered film, leaving the trajectory open where the screenplay does.
Frame-level computer vision metrics serve as unscored diagnostics for Reconstruction.
The protocol proceeds in four steps: the evaluator executes the submission (\cref{sec:evaluation-setup}), observes its objects (\cref{sec:structured-evidence}), assesses each criterion (\cref{sec:outcome-assessment}) and aggregates the results into scores (\cref{sec:score-reporting}).

\subsubsection{Submission Execution}
\label{sec:evaluation-setup}
The evaluator executes the submitted source through its prescribed entrypoint in a pinned environment. In Reconstruction, probes record object states, events and trajectories from the running project without exporting a video by default. In Creation, the evaluator renders the project and extracts clips for the judge.
Validity checks enforce the submission interface and execution rules. The submitted video can support diagnosis but does not supply the primary score.
In 3D Camerawork, a code gate also rejects hard-coded per-frame camera tables and caps the number of keyframes, so the camera must be programmed rather than traced.
A broken required interface that prevents unambiguous observation invalidates the submission.

\subsubsection{Object-Grounded Observation}
\label{sec:structured-evidence}
An \emph{object} is a production entity with an identity and a temporal scope, such as a title, a supplied image, an articulated subject or a camera. Reconstruction observes these objects in the executed project, while Creation writes its rubric questions about the objects in each shot and has the judge answer them from rendered clips (\cref{sec:outcome-assessment}).
An object's boundaries follow the production requirement rather than the code: one object may be drawn by several nodes, and an instanced collection may contain several required objects.
Motion Fusion takes a coordinated motion unit as the default object and declares subparts only when their independent editability or relative motion matters.
In 3D Camerawork, the object is the camera's state over time, including its pose, field of view, shot boundaries and motion, assessed against the known scene's geometry and subjects.

\paragraph{Public object contract.}
A JSON object list in the task inputs assigns each object a stable ID and records any prescribed code name, required parts and associated assets.
Where supplied, masks or outlines on selected reference frames use the same IDs to mark visible boundaries, as in \cref{fig:anchors}.
The agent binds these IDs to the content its project draws, which lets the evaluator locate the same object in any implementation.
Reference values, tolerances and weights remain private.

\begin{figure}[!t]
\centering
\includegraphics[width=\linewidth]{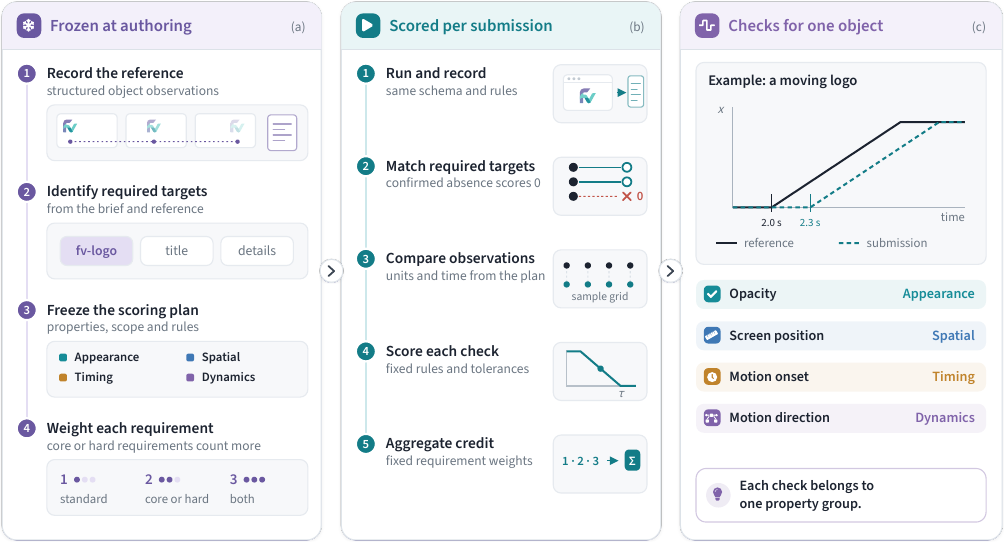}
\caption{\textbf{Object-grounded Reconstruction scoring.} (a) Authors record the reference observations and fix the targets, comparison rules and requirement weights. (b) The evaluator observes the submitted project using the same definitions, matches targets and scores the applicable checks. The task score is a weighted mean of requirement credits. (c) Checks for four properties of one object.}
\label{fig:object-evaluation-contract}
\end{figure}

\paragraph{Observation contract.}
Before evaluating submissions, authors record reference observations and define the checks through three artifacts (\cref{fig:object-evaluation-contract}a).
The \emph{observation schema} specifies which objects and fields are recorded.
The \emph{property dictionary} defines each field's unit, coordinate system, time origin and representation, so a world-space, a screen-space and a camera-relative position remain distinct channels even when each is stored as a vector.
The \emph{scoring plan} specifies the scored requirements, sampling windows, normalization, error bounds, weights and prerequisites.
The task specification and reference video decide which properties are compared, and the GT project supplies their values.
Choices the task leaves open, such as internal geometry or code organization, carry no weight merely because they appear in the GT project.

\paragraph{Measurement.}
The evaluator executes both projects and normalizes their observations into the same structure, as shown in \cref{fig:object-evaluation-contract}b.
Probes resolve each binding on the rendered nodes, meshes or groups and record the values the project actually executes, so a matching variable name or a plausible source formula does not count as evidence.
Internal names, node order and rotation representations are left to the agent, whereas public IDs and prescribed code names are fixed.
A binding or property that cannot be measured is recorded as unknown rather than inferred.
The scoring plan fixes how observed elements are matched to targets, using declared bindings where supported.
\Cref{app:structured-scoring} specifies observation packets, comparators and adapters.

\subsubsection{Output Assessment}
\label{sec:outcome-assessment}
\label{sec:reconstruction-assessment}
\label{sec:creation-assessment}
Assessment proceeds in three steps, summarized in \cref{tab:property-groups}. A presence gate first checks that each required target exists, property groups then organize the target's criteria, and each criterion is finally decided by comparison with the reference in Reconstruction or by a VLM judge in Creation.

\paragraph{Presence Gate.}
If a required target is confirmed missing, all its criteria score zero and their weights stay in the denominator. If its presence or a required measurement cannot be resolved, the task score is not reported rather than guessed.
Partial absence, intended occlusion, mistiming and attribute errors fall to the corresponding property criteria instead; for a camera, poor framing or an off-screen subject is likewise a property error.
The gate itself contributes no points.

\afterpage{%
\begin{table}[!t]
\centering\small
\setlength{\tabcolsep}{6pt}
\renewcommand{\arraystretch}{1.40}
\caption{\textbf{Shared assessment components.} A per-target presence prerequisite precedes four property groups. The assessment basis may be a GT comparison or a rubric judgment, according to the task. Each task specifies its applicable checks. Reconstruction frame comparisons provide auxiliary diagnostics. The strip below marks the four groups on the FV logo of the worked example: its position and size (Spatial), mark and colors (Appearance), rightward motion and fade (Dynamics) and the 2.0 to 3.0\,s interval (Timing); the badge marks the presence check.}
\label{tab:measurement-summary}
\label{tab:property-groups}
\resizebox{\linewidth}{!}{%
\begin{tabular}{lll}
\toprule
\tabhead{Component} & \tabhead{Assessment focus} & \tabhead{Assessment basis} \\
\midrule
\rowcolor{fvBrandViolet!8}
\assessmentlabel{fvPropGate}{eye}{Presence Gate} & Required target identity and presence within scope & GT correspondence or content rubric \\
\midrule
\rowcolor{fvBrandCyan!8}
\assessmentlabel{fvPropAppearance}{palette}{Appearance} & Shape, color, material, text, opacity, lighting, sharpness, legibility & GT attributes or judged clips \\
\rowcolor{fvBrandCyan!8}
\assessmentlabel{fvPropSpatial}{vector-square}{Spatial} & Position, size, orientation, layering, occlusion, relations, framing & GT geometry, object tracks or judged layouts \\
\rowcolor{fvBrandCyan!8}
\assessmentlabel{fvPropTiming}{clock}{Timing} & Onset, offset, duration, event order, synchronization, cuts & Event intervals, cuts or judged continuous clips \\
\rowcolor{fvBrandCyan!8}
\assessmentlabel{fvPropDynamics}{bezier-curve}{Dynamics} & Motion paths, speed, easing, rotation, deformation, attribute changes & GT trajectories, camera traces or judged clips \\
\bottomrule
\end{tabular}%
}
\par\vspace{6pt}
\includegraphics[width=\linewidth]{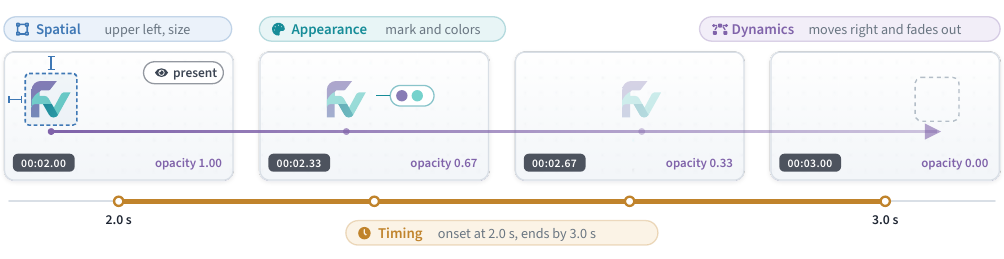}
\end{table}
}

\Needspace{4\baselineskip}
\paragraph{Property Groups.}
Appearance, Spatial, Timing and Dynamics organize the checks of every task, although a task need not cover all four.
Requirements on relations, events or the film's shared structure fall into the same groups.
Examples include contact between two parts, a child following its parent, a title disappearing before a cut and a camera revealing the required subject; each declares its endpoints and temporal scope.
Every atomic criterion is counted once.

\paragraph{Criterion Decisions.}
The two tracks decide criteria differently.
\emph{Reconstruction} compares the observed properties with the reference under the task's measurement rules. Depending on the requirement, a check compares sample values, events or motion over a specified interval (\cref{fig:object-evaluation-contract}c).
Frame diagnostics compare the evaluator's render with reference frames on the declared timeline and do not enter the score.

\emph{Creation} relies on judgment: each screenplay-derived criterion has a target, a temporal scope and a fixed pass condition.
The judge views continuous clips of the rendered film at original speed, grouped by shot and rubric aspect. Evidence starts at the screenplay timestamps. If a required action is not visible there, the judge can inspect nearby footage to distinguish a timing error from a missing action.
The judge returns \emph{pass} or \emph{fail} for each criterion.
Any design that satisfies the rubric passes, since the reference shows only one feasible solution.

\paragraph{Worked Example.}
\label{sec:assessment-example}
Consider an illustrative brief in which the FV logo appears at the upper left at 2\,s, then moves right and fades out by 3\,s; the strip under \cref{tab:property-groups} marks each step.
The presence gate first checks that the logo is drawn; if it is omitted, all of its criteria score zero while the other targets remain assessable.
Property groups then organize the logo's criteria: Appearance checks its mark and colors, Spatial its initial position and size, Timing the onset and the one-second interval, and Dynamics the rightward movement and the fade.
In Reconstruction, the evaluator compares the observed colors, position, onset and motion with the reference under their respective rules. In Creation, the judge answers separate binary questions about the same requirements from clips covering the shot.

\Needspace{6\baselineskip}
\subsubsection{Score Reporting}
\label{sec:score-reporting}
Scores are built in three steps: each criterion converts a discrepancy into credit, each task averages the credit of its criteria with fixed weights, and the benchmark averages task scores within and across categories.

\paragraph{Criterion Credit.}
Each criterion yields credit $s_j$ between 0 and 100. A numerical rule specifies where full credit ends and where credit reaches zero; between those bounds, credit can fall linearly. For example, if a position check gives full credit at zero error and none at 24\,px, a 12\,px offset earns 50. Categorical checks can use exact equality after normalization, and Creation judgments give 100 or 0. \Cref{app:structured-scoring} gives the general credit function and other comparison rules.

\paragraph{Requirement Weights.}
Each criterion belongs to one property group. In Reconstruction, a requirement $r$ has weight $w_r=1$ by default, 2 if it is core to the reference or hard to produce, and 3 if it is both. Its criteria share the weight equally, so each receives $a_j=w_r/|r|$.
Authors mark core and difficult requirements from the reference before inspecting any submission.
In Creation, each property group carries the same total weight, shared equally among its criteria.

\Needspace{8\baselineskip}
\paragraph{Task Score.}
Let $\mathcal J$ be the fixed set of required criteria of task $t$, and let the presence prerequisite $g_j\in\{0,1\}$ be zero when a target that criterion $j$ depends on is confirmed missing. The task score is the weighted mean of criterion credits:
\begin{equation}
S_t=\frac{\sum_{j\in\mathcal J}a_j\,g_j\,s_j}{\sum_{j\in\mathcal J}a_j}.
\label{eq:target-accounting}
\end{equation}
The denominator is fixed by the scoring plan, so a missing target lowers the score instead of shrinking the set of criteria. In Creation, the equal group weights make $S_t$ the mean of the four group pass rates.
Property-group scores are weighted means over the criteria of each group. \Cref{app:structured-scoring} covers relational prerequisites, per-criterion loss accounting and bounded deductions for extra behavior.

\paragraph{Benchmark Aggregation.}
Each category $c$ contains a fixed set $\mathcal T_c$ of 20 tasks with task scores $S_t$. The category and overall scores are
\begin{equation}
S_c=\frac{1}{20}\sum_{t\in\mathcal T_c}S_t,
\qquad
S_{\mathrm{overall}}=\frac{1}{5}\sum_{c=1}^{5}S_c.
\label{eq:benchmark-aggregation}
\end{equation}
Each production category thus carries equal weight, and the overall score equals the mean over all 100 tasks, with Reconstruction contributing 80\% and Creation 20\%.
Category results accompany the overall score to expose differences in production capability, and Expressive Quality, left to the planned human study, does not enter the score.
An invalid submission receives zero and remains in the aggregate, and category and overall means are reported only when every task in their set is resolved.
\Cref{app:protocols,app:repro} document the full measurement definitions and execution settings.
\FloatBarrier

\Needspace{10\baselineskip}
\subsection{Task Authoring}
\label{sec:construction}
Authoring turns an existing film or an expert brief into a task whose requirements, targets and scoring plan are fixed before any agent is evaluated.
As shown in \cref{fig:pipelines}, it proceeds in five stages. The stages iterate: reference construction and verifier development inform each other, and validation returns unresolved findings to earlier stages.
Either origin can serve either track, and \cref{app:construction-details} gives category-specific procedures.

\paragraph{(a) Specification Design.}
Authors choose a film or develop a brief based on its intended use, the available material and what the collection still needs to cover. They specify the skill under test and fix the required content, relations and timing. Experts can prescribe composition, camera behavior or coupled motion in authored briefs. The specification lists the public inputs, execution conditions, deliverables and choices left to the agent. Object-based tasks include public evidence of target identity and boundaries.

\paragraph{(b) Reference Construction.}
For an existing film, authors reconstruct an editable project and refine it by rendering intermediate versions and comparing them with the original. For an authored brief, coding agents implement the script or scene goal and experts guide revisions after viewing the renders.
Both paths produce an editable reference project and reviewed video for subsequent measurement.

\paragraph{(c) Verifier Development.}
This stage instantiates the protocol of \cref{sec:verification} for one task. Authors identify the assessment targets, decompose each requirement into criteria and choose the correspondence rule that matches targets in a submission.
For Reconstruction, GT projects, annotations or known scenes supply the required objects, events and cameras with their reference attributes and trajectories, and authors publish the object contract that locates them in submissions.
For Creation, the specification supplies binary rubric items with observable obligations, permitted variation and explicit pass conditions.
Each check is assigned once, and its applicability, weight and observation scope are fixed before candidate evaluation.
These decisions form the task's scoring plan.

\paragraph{(d) Package Assembly.}
Authors package the instructions, available assets, starter project, pinned runtime and submission interface. The reference solution, grading data and verifier internals remain private.

\paragraph{(e) Quality Validation.}
Validation asks whether the reference fulfills the specification, whether the inputs suffice, whether the verifier responds to relevant departures and whether execution is stable.
Null or frozen controls, targeted faults and valid alternatives probe missing content, temporal coverage, attribute sensitivity and aggregation.
Model trials expose ambiguities and shortcuts, repeated execution tests stability, and visual review examines the production quality of the reference and the demands of the task.
A versioned validation record documents the checks and any outstanding findings for the task version.

\FloatBarrier
\subsection{Dataset Profile}
\label{sec:overview-scale}
Each category in \cref{tab:production-capabilities} contains 20 tasks, giving 80 Reconstruction and 20 Creation tasks.
Four tasks per category (20 in total) are public, and the remaining 80 form a private held-out set.
All 100 tasks have completed quality validation, and each carries its validation record.

\begin{figure}[!t]
\centering
\includegraphics[width=\linewidth]{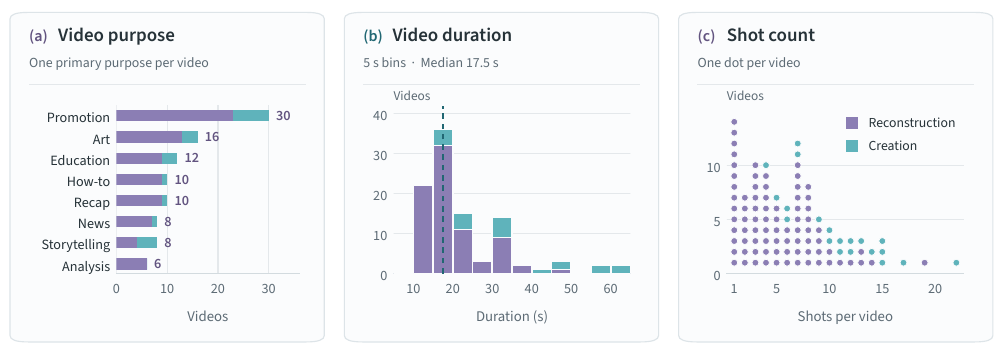}
\caption{\textbf{Video characteristics of the 100 tasks.} (a) Primary video purpose. (b) Duration in 5-second bins; the dashed line marks the median. (c) Shot count, using shot specifications for 40 tasks and video cut detection for the other 60. Colors separate Reconstruction and Creation.}
\label{fig:inventory}
\end{figure}

Selection uses three inclusion criteria: a professionally finished reference, public information sufficient to identify and carry out the required work, and an evaluation that captures the intended outcomes while accepting valid implementation differences.
Among eligible tasks, we seek variation in source families, production styles and difficulty within each category, guided by demands such as dense spatial arrangements, temporal dependencies and coupled motion. Tasks derived from the same source or from closely related variants are kept apart.
\Cref{fig:inventory} summarizes the 100 videos by primary purpose, duration and shot count.

\section{Experiments}
\label{sec:experiments}
\suppressfloats[t]

We first compare seven complete agent systems across the five categories (\cref{sec:trials}) and then use the object-level score records to diagnose where and how they fail (\cref{sec:diagnostic-analysis}). Paired stills illustrate typical failures (\cref{sec:failure-analysis}), and measurement controls delimit what the primary score can detect (\cref{sec:validation}).

\subsection{Experimental Setup}
\label{sec:experiment-setup}
\paragraph{Agent Systems.}
We evaluate seven systems built on Claude Opus~5.5~\citep{anthropic2026opus55}, GPT-6.1 Sol~\citep{openai2026gpt61sol}, GLM~5.3~\citep{glm2026glm5}, DeepSeek~V4.1 Flash~\citep{deepseek2026v41flash}, MiMo V2.6 Pro~\citep{xiaomi2026mimov26}, Kimi K3~\citep{kimi2026k3} and Step 5 Preview~\citep{stepfun2026step5}. GPT-6.1 Sol runs in Codex~\citep{openai2026codex} and the others in Claude Code~\citep{anthropic2026claudecode}. Each entry pairs a model with a harness, tools, reasoning settings and a feedback loop that stay fixed across tasks. Because systems can differ in all of these components, the comparison estimates how well each configuration performs as a whole. Run records retain the exact model identifiers, harness versions and visual-feedback interfaces.

\paragraph{Execution Budget.}
Each system has six hours per task for planning, coding, tool use, rendering and inspection. All systems receive the same public inputs and runtime for a given task and follow the tool and asset rules in \cref{sec:task-formulation}. The task container has no internet access apart from the model endpoint, and web search is disabled. Containers provide 4 CPUs and 8\,GB of memory without GPUs; 3D Camerawork tasks receive twice these resources. Verification takes place after submission, outside the agent's budget.

\paragraph{Evaluation Set.}
Each system makes one attempt on each of the 20 tasks per category. Reconstruction scores are offline regrades of the archived submissions under their recorded task scoring plans.

\subsection{Main Results}
\label{sec:trials}
\Cref{tab:controlled-results} reports the scores of the seven systems. Each category mean averages 20 task scores, and Overall follows \cref{eq:benchmark-aggregation}. The Reconstruction mean below is the unweighted mean of its four categories.

\begin{table}[!t]
\centering\small
\renewcommand{\arraystretch}{1.42}
\caption{\textbf{Category and overall scores of seven agent systems.} The harness column identifies Claude Code or Codex. Visual, Footage, Motion and Camera each average 20 tasks. Script averages the 20 tasks of the Script Realization study (\cref{app:script-scoring}). Overall is the unweighted mean of the five categories. Ranks follow Overall, and shaded bold values mark the best in each column.}
\label{tab:controlled-results}
\setlength{\tabcolsep}{7pt}
\begin{tabular}{@{}l c l r r r r r@{\hspace{7pt}}}
\toprule
 & & & \multicolumn{4}{c}{\tabhead{Reconstruction}} & \multicolumn{1}{c}{\tabhead{Creation}} \\
\cmidrule(lr){4-7}\cmidrule(l){8-8}
\tabhead{Model} & \tabhead{Harness} & \tabhead{Overall} & \tabhead{Visual} & \tabhead{Footage} & \tabhead{Motion} & \tabhead{Camera} & \tabhead{Script} \\
\midrule
\fvrank{1}\raisebox{-0.2em}{\includegraphics[height=1.05em]{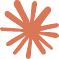}}\hspace{5pt}Claude Opus~5.5 & \raisebox{-0.2em}{\includegraphics[height=1.05em]{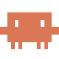}} & \fvscorebar{60.9}{\textbf{60.9}} & \fvbest{72.9} & \fvbest{69.0} & \fvbest{26.4} & 60.7 & \fvbest{75.3} \\
\fvrank{2}\raisebox{-0.2em}{\includegraphics[height=1.05em]{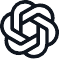}}\hspace{5pt}GPT-6.1 Sol & \raisebox{-0.2em}{\includegraphics[height=1.05em]{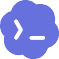}} & \fvscorebar{51.4}{51.4} & 55.9 & 45.1 & 21.7 & \fvbest{62.5} & 71.7 \\
\fvrank{3}\raisebox{-0.2em}{\includegraphics[height=1.05em]{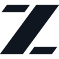}}\hspace{5pt}GLM 5.3 & \raisebox{-0.2em}{\includegraphics[height=1.05em]{assets/model-logo-claudecode.pdf}} & \fvscorebar{38.7}{38.7} & 49.3 & 50.9 & 17.9 & 21.7 & 53.7 \\
\fvrank{4}\raisebox{-0.2em}{\includegraphics[height=1.05em]{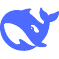}}\hspace{5pt}DeepSeek~V4.1 Flash & \raisebox{-0.2em}{\includegraphics[height=1.05em]{assets/model-logo-claudecode.pdf}} & \fvscorebar{29.8}{29.8} & 35.6 & 14.5 & 13.3 & 21.7 & 63.9 \\
\fvrank{5}\raisebox{-0.2em}{\includegraphics[height=1.05em]{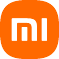}}\hspace{5pt}MiMo V2.6 Pro & \raisebox{-0.2em}{\includegraphics[height=1.05em]{assets/model-logo-claudecode.pdf}} & \fvscorebar{29.0}{29.0} & 31.6 & 21.3 & 14.1 & 18.5 & 59.6 \\
\fvrank{6}\raisebox{-0.2em}{\includegraphics[height=1.05em]{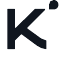}}\hspace{5pt}Kimi K3 & \raisebox{-0.2em}{\includegraphics[height=1.05em]{assets/model-logo-claudecode.pdf}} & \fvscorebar{27.8}{27.8} & 32.4 & 18.1 & 14.7 & 16.0 & 58.0 \\
\fvrank{7}\raisebox{-0.2em}{\includegraphics[height=1.05em]{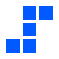}}\hspace{5pt}Step 5 Preview & \raisebox{-0.2em}{\includegraphics[height=1.05em]{assets/model-logo-claudecode.pdf}} & \fvscorebar{26.8}{26.8} & 34.1 & 19.8 & 12.9 & 22.0 & 45.2 \\
\bottomrule
\end{tabular}
\end{table}

\paragraph{Overall Ranking.}
Claude Opus~5.5 leads Overall with 60.9, followed by GPT-6.1 Sol at 51.4 and GLM~5.3 at 38.7; the remaining four systems score between 26.8 and 29.8. We resample tasks within each Reconstruction category and hold Script Realization scores fixed. The first three positions are unchanged in all 10{,}000 bootstrap draws. The intervals for the last four systems overlap, making their relative order less stable on this task set.

\paragraph{Category Profile.}
On Reconstruction alone, Opus~5.5 averages 57.2 and GPT-6.1 Sol 46.3. The two leading systems divide the categories: Opus~5.5 leads Visual Recreation, Footage Editing and Motion Fusion, while GPT-6.1 Sol leads 3D Camerawork with 62.5 against 60.7, a category in which no other system exceeds 23. Motion Fusion has the lowest category mean for every system, and none exceeds 26.4.

\paragraph{Track Comparison.}
Script Realization scores range from 45.2 to 75.3, and each system scores higher on Script than on its Reconstruction mean. The orders also differ: DeepSeek~V4.1 Flash ranks third on Script with 63.9 but sixth on Reconstruction, whereas Step 5 Preview ranks fourth on Reconstruction and last on Script. The Spearman rank correlation between the two tracks is 0.43 across the seven systems.

\subsection{Diagnostic Analysis}
\label{sec:diagnostic-analysis}
The score records show which targets and properties lose credit and which tasks remain hard.

\begin{figure}[!t]
\centering
\includegraphics[width=\linewidth]{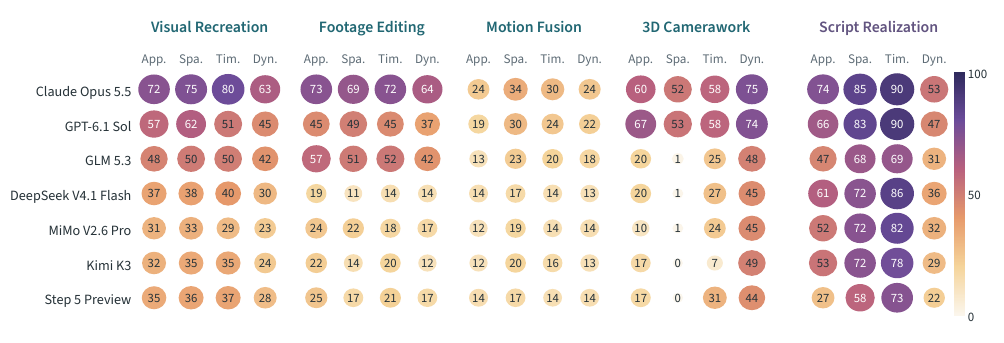}
\caption{\textbf{Property-group scores by category.} Each bubble averages one system's scores in a property group over the 20 tasks of a category, with area and color both encoding the score, and Script Realization bubbles give group pass rates. Reconstruction bubbles use the 543 of 560 runs whose score records keep the group breakdown, and a run scored zero counts as zero in every group. Systems are ordered by Overall.}
\label{fig:property-groups}
\end{figure}

\paragraph{Property Groups.}
\Cref{fig:property-groups} breaks the scores down by category and property group. In 3D Camerawork, the five lower-ranked systems score 0.2 to 1.2 on Spatial and 44 to 49 on Dynamics. Claude Opus~5.5 and GPT-6.1 Sol score 52 to 53 on Spatial. The lower-ranked systems thus reproduce roughly how the camera moves but not what it frames. In Motion Fusion, every group's score is below 34 for every system, so the difficulty is not confined to one property. In Script Realization, all seven systems score highest on Timing and lowest on Dynamics, which ranges from 22.1 to 52.6: scripted timing is met far more often than scripted motion.

\begin{figure}[!t]
\centering
\includegraphics[width=\linewidth]{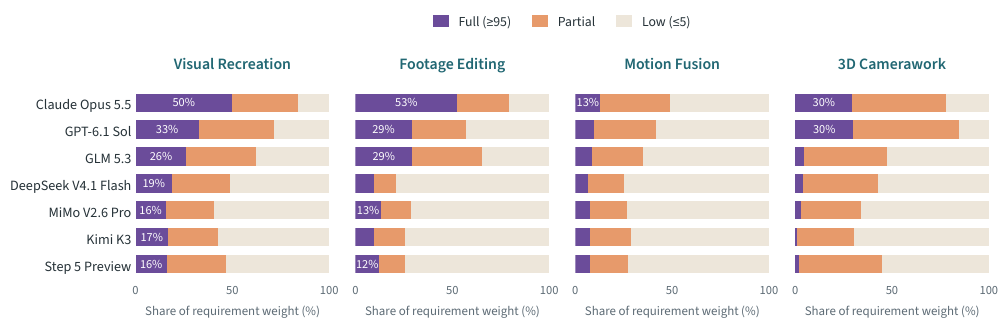}
\caption{\textbf{Requirement credit by category.} Each bar divides requirement weight into scores of at least 95 (full), at most 5 (low), and the range in between (partial), averaged across available runs for each system and category. Labels show the full-credit share. The analysis uses 510 of 560 runs with requirement-level records or a gate score of zero. In gate-zero runs all requirements enter the lowest band. Systems are ordered by Overall.}
\label{fig:requirement-credit}
\end{figure}

\paragraph{Requirement Credit.}
\Cref{fig:requirement-credit} groups requirement credits into full (at least 95), partial and low (at most 5). The respective shares of requirement weight are 17\%, 29\% and 54\%. Claude Opus~5.5 has 37\% in the full-credit group and 36\% in the partial-credit group, while the four lowest-ranked systems have 9 to 10\% in the full-credit group. Most requirement weight therefore earns little or no credit, and the leading system stands out mainly by satisfying more requirements completely.

\paragraph{Object-Level Loss.}
Targets that earn no credit on any requirement account for 80\% of the lost points in Footage Editing and 52\% in Motion Fusion. The corresponding shares are 19\% in Visual Recreation and 7\% in 3D Camerawork. Footage Editing and Motion Fusion therefore lose most credit on targets that are missed entirely, whereas Visual Recreation and 3D Camerawork lose it mainly on targets that are present but imprecise. The calculation includes runs assigned zero by a submission gate.

\paragraph{Task Difficulty.}
Within each category, the Spearman correlation between a task's mean score and its duration or shot count lies within $\pm0.24$. In Footage Editing, the correlation with object count is $-0.40$. Difficulty thus depends on what a task asks for more than on its length. No system reaches 50 on 26 of the 80 Reconstruction tasks, including all 20 Motion Fusion tasks, while at least one system exceeds 95 on 6 tasks. Opus~5.5 attains the highest score on 60 tasks and GPT-6.1 Sol on 16, nine of them in 3D Camerawork.

\FloatBarrier
\Needspace{12\baselineskip}
\subsection{Qualitative Failures}
\label{sec:failure-analysis}
\begin{figure}[!t]
\centering
\newcommand{\failurepanel}[5][]{%
\begin{minipage}[t]{.485\linewidth}
\footnotesize\raggedright
{\sffamily\textcolor{fvBrandTeal}{\textbf{#2}}}\par\vspace{2pt}
{\scriptsize\sffamily\color{metagray}%
\makebox[.486\linewidth]{Reference}\hfill\makebox[.486\linewidth]{Agent output}}\par
\noindent #3\par\vspace{2pt}
\textbf{#4.} #5
\end{minipage}}
\failurepanel{(a) Visual Recreation}{\includegraphics[width=.486\linewidth]{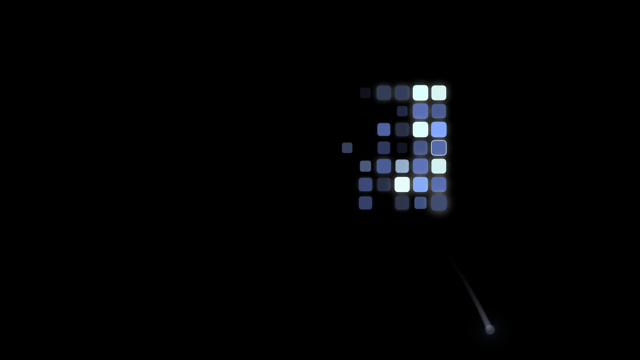}\hfill\includegraphics[width=.486\linewidth]{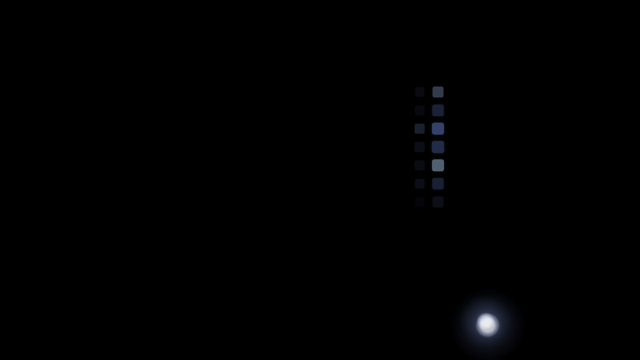}}{Aureole, 23.28\,s}{The reference assembles a grid of cells; the output draws only a faint column.}\hfill
\failurepanel{(b) Footage Editing}{\includegraphics[width=.486\linewidth]{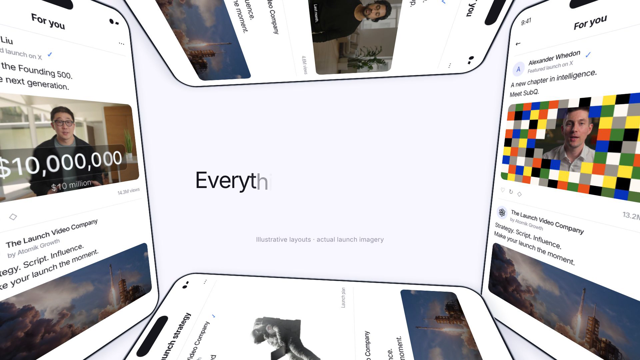}\hfill\includegraphics[width=.486\linewidth]{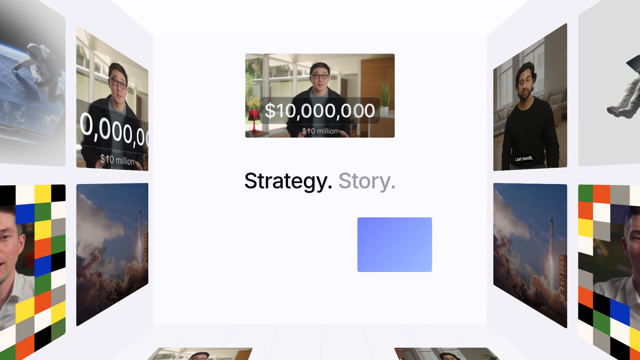}}{Astra, 8.95\,s}{The same media appear as flat cards instead of tilted phone screens.}\par\vspace{7pt}
\failurepanel{(c) Motion Fusion}{\includegraphics[width=.486\linewidth]{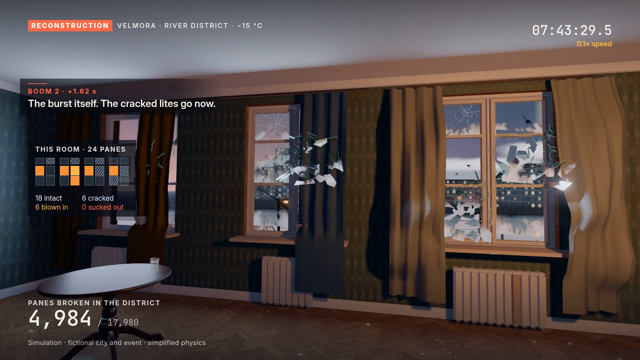}\hfill\includegraphics[width=.486\linewidth]{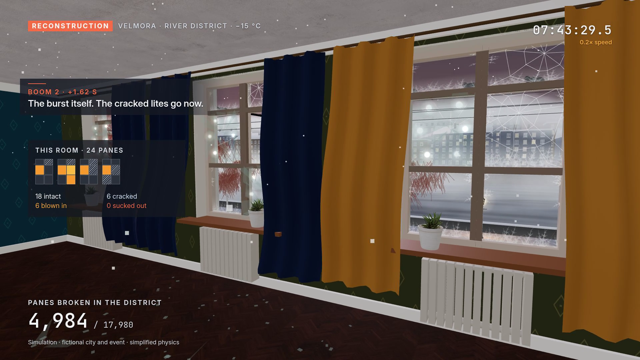}}{Fireball, 8.91\,s}{The reference shows glass bursting into the room; the output's windows stay intact.}\hfill
\failurepanel{(d) 3D Camerawork}{\includegraphics[width=.486\linewidth]{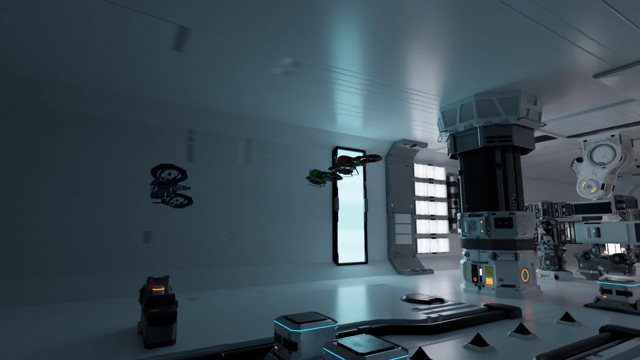}\hfill\includegraphics[width=.486\linewidth]{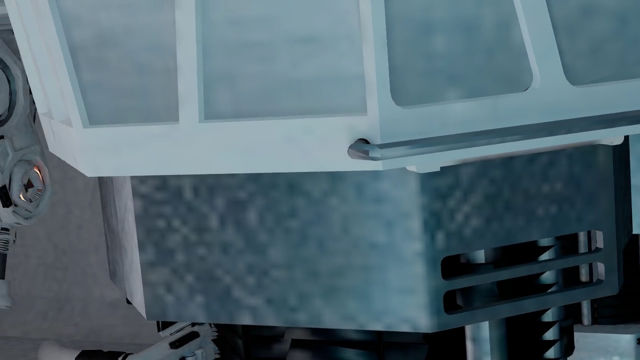}}{Zero-Day Drone Race, 8.36\,s}{The reference frames the drones in the hall; the output camera faces nearby machinery.}
\caption{\textbf{Qualitative failures across four Reconstruction categories.} Paired stills at matching times compare a reference with the delivered video of the Claude Opus~5.5 run in the controlled comparison. The examples illustrate distinct errors, not their frequency.}
\label{fig:reconstruction-failures}
\end{figure}

The aggregate patterns have counterparts in individual films. \Cref{fig:reconstruction-failures} shows requirements that fail even though the intended content is recognizable: a visual detail is missing, familiar media are arranged differently, an event is not visible at the expected time, or the camera misses the subject.

\paragraph{Camera Framing.}
Panel (d) is an instance of the Spatial gap in 3D Camerawork shown in \cref{fig:property-groups}: the output camera is in the right scene, but its pose leaves the drones out of frame. Cut times earn Timing credit, whereas framing within each shot depends on the poses, which Spatial and Dynamics score separately.

\paragraph{Coupled Motion.}
Panel (c) shows the kind of event that Motion Fusion requires. A single still can localize the discrepancy, but establishing when the glass breaks, or whether it breaks at all, requires evidence over time.

\subsection{Measurement Validation}
\label{sec:validation}
Measurement controls establish what the primary score can and cannot detect. Scoring the reference execution checks that the scoring rules recover the target values, and comparing the reference render with the input video checks visual fidelity. Null and frozen controls test sensitivity to missing content and motion, and repeated executions test stability. Validation also confirms that each declared object ID resolves to what the project actually draws and that properties are compared over the required temporal scope, so naming the correct object without rendering it earns no credit (\cref{app:validation-details}). \Cref{app:earlier-trials} records how earlier protocol versions overstated fidelity when their measurements missed part of the film.

\paragraph{Framing Hidden by a High Total.}
An offline control on the Apollo 11 Motion Fusion task doubled the tangent of the camera's half field of view in the main view of the reference, holding world geometry and secondary views fixed. Relative to the identity control, the total score fell only from 100 to 95.46, while the framing subset fell from 100 to 21.94 under the same scoring plan. The framing subset reports the original positive factors for camera state and planned screen position and size and adds no weight to the total. Native renders at five fixed times confirm the framing discrepancy. Reporting such subsets alongside the total exposes errors that a weighted mean dilutes.
\FloatBarrier

\section{Discussion and Limitations}
\label{sec:lessons}

A score is meaningful only together with its measurement scope. The controls in \cref{sec:validation,app:earlier-trials} show that a submission can score highly while omitting visible content or freezing every object, so the evaluator checks that each measurement follows the content actually drawn and reports which requirements it covers.

\paragraph{Coverage.}
The 20 tasks per category give each of the five production demands equal representation. Difficulty can still differ across categories, and the scope extends only to the production work represented by the selected briefs and reference films. Footage Editing currently composes still images, vector graphics and fonts; cutting supplied video clips is not yet covered. Audio, originality and ease of revision are not scored. Executable source enables intervention tests of editability, but does not establish that revisions will be easy or correct. A stated style requirement, such as a prescribed typeface, can be scored directly, whereas whole-film qualities such as coherence across shots require separate judgments (\cref{sec:future-work}).

\paragraph{Measurement Validity.}
Structured records compare different implementations only when their measurements share a definition and trace to execution, which the quality validation of \cref{sec:construction} tests for each task. A changed measurement or comparison rule requires revalidation under a new version. Coverage also depends on which objects a plan scores: some current plans leave backgrounds unscored or overweight static relations.

\paragraph{Experimental Scope.}
The comparison uses one six-hour attempt per task. The bootstrap describes variation within the Reconstruction task set, conditional on the observed Script scores; it does not estimate the variability of repeated agent runs. The design does not isolate the effects of budget or feedback. A failed output alone cannot tell us whether the agent misunderstood the task or implemented it incorrectly. Agents run without internet access, and the 80 held-out tasks keep their inputs, references and grading data private. Some references derive from publicly available projects, however, and a model may have seen that code during training.

\paragraph{Generality and Cost.}
The object-grounded protocol is not tied to one renderer. The observation schema, property dictionary and scoring plan are shared, and only the probe that reads object states differs across the Remotion projects, three.js scene descriptions and Blender camera programs used here. Extending it to UI motion would mainly require a probe for the target UI framework, since animated interface elements expose identities and properties much as DOM nodes do. Games are a similar case: engines already organize scenes into entities with identities, transforms and animation states, so an engine-side probe would allow the same object-level evaluation. In every domain, the larger cost is likely to remain the construction and validation of an executable reference project for each task.

\section{Future Work}
\label{sec:future-work}

Several parts of the evaluation are specified but have not yet been run.

\paragraph{Human Assessment.}
Two human studies will complement the automatic scores. The first quantifies the reliability of the Creation VLM judge, whose errors currently enter Script Realization scores unmeasured. Human experts judge the same screenplay criteria from the same evidence, and the study reports criterion-level agreement with the judge (Cohen's $\kappa$), false-pass and false-rejection rates, the rank correlation of task scores with human scores (Spearman's $\rho$) and the stability of repeated judge runs. The second rates whole-film Expressive Quality across all five categories as a secondary outcome that does not enter the benchmark score. \Cref{sec:human-studies} specifies the sampling, blinding and measurements of both studies.

\paragraph{Measurement Extensions.}
The frame-level diagnostics in \cref{fig:assessment-mechanisms}c do not yet include semantic similarity. The framing subset in \cref{sec:validation} exposes errors that the Motion total dilutes, but it has not yet been calibrated against human judgments or validated on held-out renders.

\section{Conclusion}
\label{sec:conclusion}

\fv evaluates coding agents on professional video production through object-grounded verification. Its tasks span four Reconstruction categories, covering the frame, the media, the motion and the camera, and one screenplay-driven Creation category, and every task asks for an editable project and a rendered film. In Reconstruction, a public object contract lets the evaluator read each object's properties from the executed project and compare them with the reference, so different implementations that satisfy the same checks earn equal credit. In Creation, a VLM judge answers screenplay-derived questions about the objects in each shot from the rendered film. A five-stage authoring pipeline fixes each task's targets and scoring plan before any agent is evaluated and validates them with controls.

Across seven agent systems, the strongest reaches 60.9 overall and none exceeds 26.4 on Motion Fusion. Footage Editing and Motion Fusion lose most credit on targets that are missed entirely, and in 3D Camerawork the lower-ranked systems reproduce how the camera moves but not what it frames. The two tracks also rank systems differently, so neither substitutes for the other. Object-level records thus show not only how much a system scores but where it loses credit.

\FloatBarrier
\Needspace{14\baselineskip}
\pdfbookmark[0]{Author Contributions}{fv-author-contributions}
\section*{Author Contributions}
\begingroup\raggedright
\renewcommand{\arraystretch}{1.12}
\newcommand{\fvcofirst}{\ensuremath{\ast}}

\paragraph{Core Team.}\mbox{}\par
\noindent
\begin{tabularx}{\linewidth}{@{}*{4}{>{\raggedright\arraybackslash}X}@{}}
\mbox{Tianfu Wang\textsuperscript{1,\fvcofirst,\textdaggerdbl}} & \mbox{Yin Wu\textsuperscript{1,\fvcofirst}} & \mbox{Yuan Feng\textsuperscript{2,\fvcofirst}} & \mbox{Zhezheng Hao\textsuperscript{3,\fvcofirst}} \\[2pt]
\mbox{Haowen Yang\textsuperscript{1,\fvcofirst}} & \mbox{Yi Zhan\textsuperscript{2,\fvcofirst}} & \mbox{Huiling Pan\textsuperscript{4}} &  \\[2pt]
\mbox{Peijie Dong\textsuperscript{1}} & \mbox{Jianqing Zhang\textsuperscript{5}} & \mbox{Lixin Liu\textsuperscript{6}} &
\end{tabularx}\par

\paragraph{Contributors.}\mbox{}\par
{\small Contributions cover benchmark data, method discussions and manuscript feedback. Contributors appear alphabetically by surname.\par}
\smallskip
\noindent
\begin{tabularx}{\linewidth}{@{}*{4}{>{\raggedright\arraybackslash}X}@{}}
\mbox{Xi Chen\textsuperscript{2}} & \mbox{Liwei Deng\textsuperscript{7}} & \mbox{Linxiao Gong\textsuperscript{1}} & \mbox{Yingxin Jia\textsuperscript{8}} \\[2pt]
\mbox{Xinhang Li\textsuperscript{2}} & \mbox{Yule Liu\textsuperscript{1}} & \mbox{Haochen Luo\textsuperscript{9}} & \mbox{Zhiheng Lv\textsuperscript{10}} \\[2pt]
\mbox{Zhuoshi Pan\textsuperscript{6}} & \mbox{Rui Qian\textsuperscript{11}} & \mbox{Hui Ren\textsuperscript{10}} & \mbox{Yichen Wan\textsuperscript{2}} \\[2pt]
\mbox{Haosen Wang\textsuperscript{19}} & \mbox{Junyang Wang\textsuperscript{2}} & \mbox{Xilin Xia\textsuperscript{2}} & \mbox{Yuanhang Xiao\textsuperscript{12}} \\[2pt]
\mbox{Yizhao Xu\textsuperscript{13}} & \mbox{Ruofeng Yang\textsuperscript{5}} & \mbox{Yongkang Yang\textsuperscript{2}} & \mbox{Xingbo Yao\textsuperscript{1}} \\[2pt]
\mbox{Jinhui Ye\textsuperscript{4}} & \mbox{Mu Yuan\textsuperscript{12}} & \mbox{Jiahao Zeng\textsuperscript{1}} & \mbox{Xichong Zhang\textsuperscript{2}} \\[2pt]
\mbox{Dongzhuoran Zhou\textsuperscript{14}} & \mbox{Yangjunfeng Zhu\textsuperscript{6}} &  & 
\end{tabularx}\par

\paragraph{Advisory Committee.}\mbox{}\par
\noindent
\begin{tabularx}{\linewidth}{@{}*{4}{>{\raggedright\arraybackslash}X}@{}}
\mbox{Hui Xiong\textsuperscript{1,\textdagger}} & \mbox{Qiang Yang\textsuperscript{15}} & \mbox{Dacheng Tao\textsuperscript{16}} & \mbox{Nicholas Jing Yuan\textsuperscript{1}} \\[2pt]
\mbox{Yi Zhu\textsuperscript{17}} & \mbox{Li Shen\textsuperscript{18}} & \mbox{Xuming Hu\textsuperscript{1}} & \mbox{Zishuo Ding\textsuperscript{1}} \\[2pt]
\mbox{Jiawei Chen\textsuperscript{3}} & \mbox{Xike Xie\textsuperscript{2}} & &
\end{tabularx}\par
\medskip
{\small
\noindent\textsuperscript{\fvcofirst}\,Co-first authors.\quad\textsuperscript{\textdaggerdbl}\,Project leader.\par
\smallskip
\noindent\textsuperscript{\textdagger}\,Advisory committee chair.\par}
\medskip
{\fontsize{9.5}{11.5}\selectfont
\noindent
\textsuperscript{1}\,The Hong Kong University of Science and Technology (Guangzhou)\\
\textsuperscript{2}\,University of Science and Technology of China\\
\textsuperscript{3}\,Zhejiang University\\
\textsuperscript{4}\,The Hong Kong University of Science and Technology\\
\textsuperscript{5}\,Shanghai Jiao Tong University\\
\textsuperscript{6}\,Tsinghua University\\
\textsuperscript{7}\,Aalborg University\\
\textsuperscript{8}\,Meta\\
\textsuperscript{9}\,The University of Hong Kong\\
\textsuperscript{10}\,University of Illinois Urbana-Champaign\\
\textsuperscript{11}\,Fudan University\\
\textsuperscript{12}\,The Chinese University of Hong Kong\\
\textsuperscript{13}\,Peking University\\
\textsuperscript{14}\,University of Oslo\\
\textsuperscript{15}\,The Hong Kong Polytechnic University\\
\textsuperscript{16}\,Nanyang Technological University\\
\textsuperscript{17}\,Prana Labs\\
\textsuperscript{18}\,Sun Yat-sen University\\
\textsuperscript{19}\,Independent Researcher\par}
\endgroup

\bibliographystyle{fv-unsrtnat}
\bibliography{references,bib/intro,bib/overview,bib/construction,bib/verification,bib/validation}

\clearpage
\beginappendix
\pdfbookmark[0]{Appendix Contents}{fv-appendix-contents}
\vspace{0.65em}
\begin{tcolorbox}[
  enhanced,colback=fvBrandViolet!9!white,colframe=fvBrandViolet!9!white,
  boxrule=0pt,arc=3pt,boxsep=0pt,left=8pt,right=8pt,top=5pt,bottom=5pt,
  before skip=0pt,after skip=0pt]
  {\sffamily\small\bfseries\color{fvBrandInkViolet}Contents\hfill
  \textcolor{fvBrandTeal}{Page}}\par
\end{tcolorbox}
\fvappendixcontents

\clearpage
\section{Benchmark Details}
\label{app:design-details}
\label{app:tasks}
\suppressfloats[t]
The task composition and submission contract are defined in \cref{sec:overview-scale,sec:task-formulation}. The public evidence below illustrates how object-based tasks make target identities and boundaries available to agents.

\subsection{Public Evidence}
\begin{figure}[!t]
\centering
\begin{minipage}[t]{0.192\linewidth}\centering
\includegraphics[width=\linewidth]{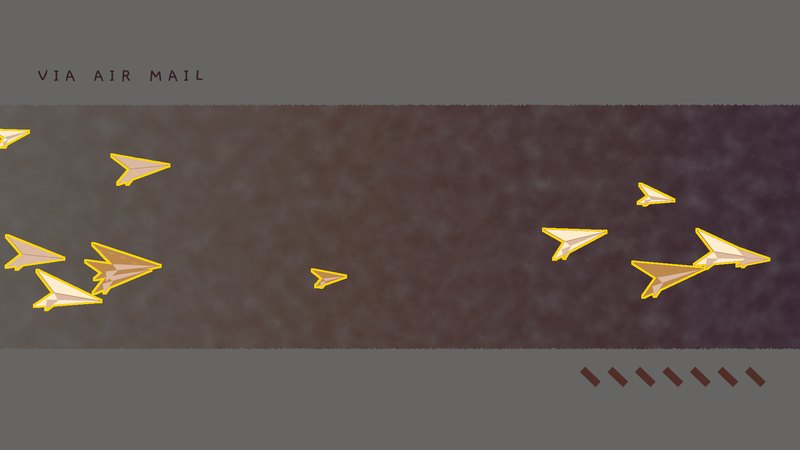}\\[1pt]
\slug{9\,s \textperiodcentered{} airmail.planes}
\end{minipage}\hfill
\begin{minipage}[t]{0.192\linewidth}\centering
\includegraphics[width=\linewidth]{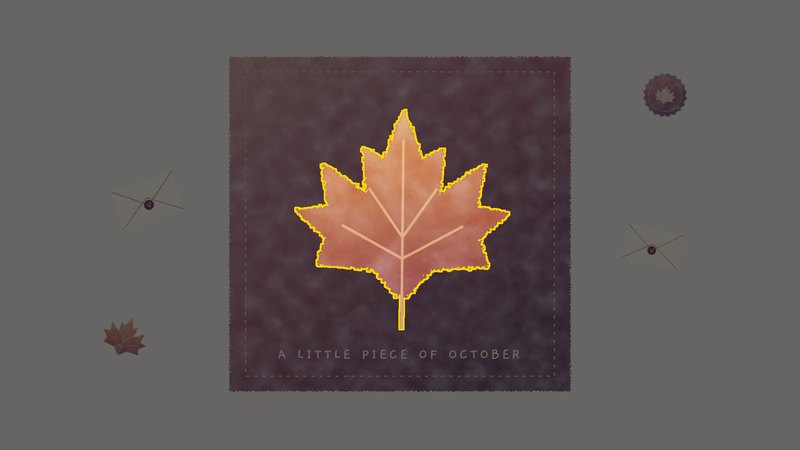}\\[1pt]
\slug{17\,s \textperiodcentered{} maple.hero}
\end{minipage}\hfill
\begin{minipage}[t]{0.192\linewidth}\centering
\includegraphics[width=\linewidth]{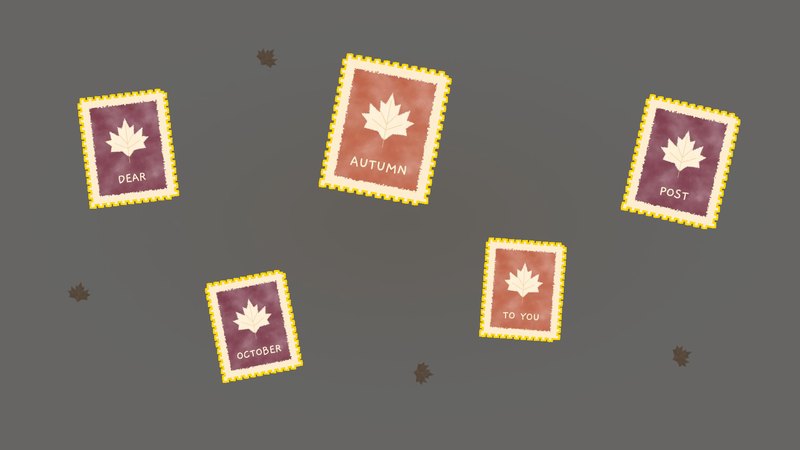}\\[1pt]
\slug{19.5\,s \textperiodcentered{} stamps.group}
\end{minipage}\hfill
\begin{minipage}[t]{0.192\linewidth}\centering
\includegraphics[width=\linewidth]{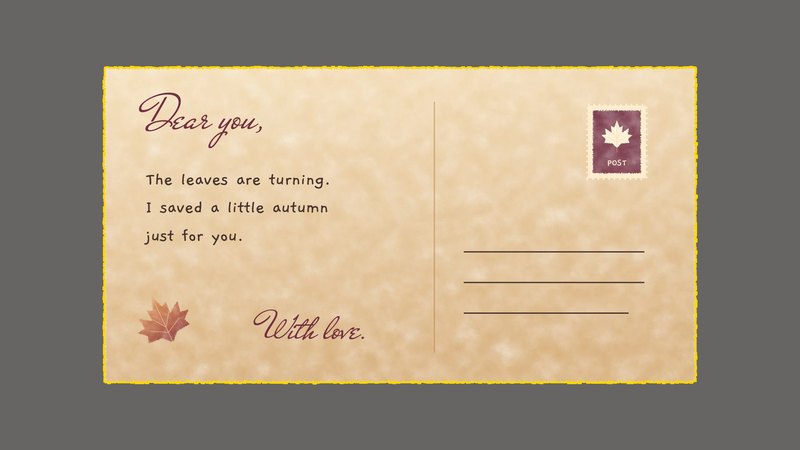}\\[1pt]
\slug{22.5\,s \textperiodcentered{} letter.card}
\end{minipage}\hfill
\begin{minipage}[t]{0.192\linewidth}\centering
\includegraphics[width=\linewidth]{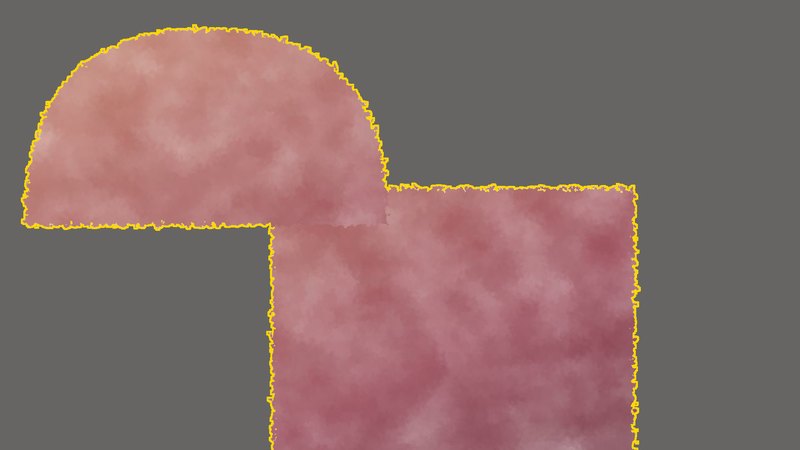}\\[1pt]
\slug{26\,s \textperiodcentered{} mailbox.assembly}
\end{minipage}
\caption{\textbf{Public object evidence for \texttt{reconstruct-visual-autumn-post}.} The five reference frames the agent receives, each with its instance mask applied: the masked object keeps its colors and is outlined, and the rest of the frame is dimmed. The five objects appear in five non-overlapping shots, so the task provides five frames instead of the default four. Masks are generated from what the trusted reference project actually draws, keeping transforms, crops, opacity and effects. The agent receives the frames and masks; the scoring rules stay hidden.}
\label{fig:anchors}
\end{figure}

\paragraph{Object Identity.}
In Visual Recreation tasks such as \texttt{autumn-post} (\cref{fig:anchors}), \texttt{objects.json} links target IDs and prescribed code names to required parts and any corresponding asset files. The generated \texttt{src/scene-objects.ts} module supplies those names and helpers for binding them to DOM/SVG elements or canvas draw calls. Three.js tasks can attach the target ID to the mesh or group through \texttt{userData.benchObject}.

\paragraph{Visible Boundaries.}
The input design calls for recognizable boundary evidence for visible objects assessed at the object level. Examples include a supplied asset linked to its target ID, a visible-region mask or an approximate outline paired with a reference frame. Several objects can share a frame, with separate layers where they overlap. Non-visual targets use public interfaces and observable effects. \Cref{fig:anchors} shows masks from a multi-shot task.

The task marks each target and property as scored, context only or deferred before evaluation. The fixed scoring plan applies that scope to the criteria and denominator. Public images identify targets. Trajectories, event answers, weights and tolerances remain private.

\subsection{Category Specifications}
\label{sec:category-specs}
\Cref{tab:category-specs} summarizes what each category supplies to the agent, what it asks the agent to deliver and what the evaluator compares against.

\begin{table}[!t]
\centering\footnotesize
\caption{\textbf{Category specifications.} Inputs beyond the reference video or screenplay, the editable deliverable, the reference used for scoring and the primary scoring evidence.}
\label{tab:category-specs}
\resizebox{\linewidth}{!}{%
\begin{tabular}{@{}lllll@{}}
\toprule
\tabhead{Category} & \tabhead{Additional inputs} & \tabhead{Deliverable} & \tabhead{Reference} & \tabhead{Primary evidence} \\
\midrule
Visual Recreation & \makecell[lt]{Object table, reference\\frames and masks} & Remotion project & \makecell[lt]{Original or rebuilt\\source project} & \makecell[lt]{Properties of\\drawn objects} \\[3pt]
Footage Editing & \makecell[lt]{Byte-identical images, vectors\\and fonts; object table} & Remotion project & \makecell[lt]{Original or rebuilt\\source project} & \makecell[lt]{Object properties, assets\\identified by content hash} \\[3pt]
Motion Fusion & \makecell[lt]{Object evidence and masks,\\or the reference video alone} & \makecell[lt]{Remotion project or\\3D scene description} & \makecell[lt]{Annotated reference\\films by the team} & \makecell[lt]{Object and event requirements,\\or scene motion fields} \\[3pt]
3D Camerawork & \makecell[lt]{Fixed 3D scene with its geometry\\and animation; rendering service} & \makecell[lt]{Blender camera\\program} & \makecell[lt]{Ground-truth\\camera trajectory} & \makecell[lt]{Camera pose, focal\\length and cut times} \\[3pt]
Script Realization & Soundtrack where provided & Remotion project & \makecell[lt]{None; the screenplay\\defines the obligations} & \makecell[lt]{Judged clips of\\the rendered film} \\
\bottomrule
\end{tabular}%
}
\end{table}

\subsection{Output Formats}
\label{sec:overview-contracts}
Reference video and output formats use native 720p or 1080p, in landscape or portrait orientation. Three Creation tasks keep their native sizes: \texttt{sweet-tooth} (1080 by 1080), \texttt{abyssal} (1008 by 720), and \texttt{this-is-math} (320 by 568). Reconstruction scores visual requirements and records audio presence as metadata.

\FloatBarrier
\section{Evaluation Details}
\label{app:protocols}
\label{sec:verif-principles}
This appendix gives the details behind \cref{sec:verification}. \Cref{tab:assessment-targets} lists a typical assessment target and observation for each category; each task's scoring plan fixes its own measured properties and coverage. Where a task requires independent editability or part-specific behavior, it declares the parts and relations to check.

\begin{table}[!t]
\centering\footnotesize
\caption{\textbf{Example targets and observations by category.} Each task's fixed scoring plan specifies what it measures and scores.}
\label{tab:assessment-targets}
\begin{tabularx}{\linewidth}{@{}l>{\raggedright\arraybackslash}p{3.4cm}>{\raggedright\arraybackslash}X@{}}
\toprule
\tabhead{Category} & \tabhead{Target} & \tabhead{Example observation or criterion} \\
\midrule
Visual Recreation & Objects and visual groups & Title content, appearance and layout \\
Footage Editing & Material instances & Asset identity, crop, layering and entry time \\
Motion Fusion & Motion units and relations & Coupled motion, contact and deformation \\
3D Camerawork & Camera state over time & Pose, projection, framing and cuts \\
Script Realization & Required entities and events & Subject presence, actions and event order \\
\bottomrule
\end{tabularx}
\end{table}

\FloatBarrier

\subsection{Property Dictionary}
\label{sec:property-dictionary}
\Cref{tab:property-dictionary} lists the principal observed channels of each Reconstruction category by property group. Every channel fixes its unit, coordinate frame and clock in the property dictionary, and every comparison uses one of a small set of metric families: normalized Euclidean distance, absolute log ratio, geodesic rotation angle, absolute difference, exact equality after canonicalization, and ordered one-to-one event matching that charges missing and extra events. A task may normalize position errors by a declared reference scale, such as the frame diagonal, an element length or the median scene depth. The full-credit and zero-credit bounds are set per comparison in the private scoring plan. They are protocol parameters, not estimates of perceptual thresholds.

\begin{table}[!t]
\centering\footnotesize
\caption{\textbf{Principal observed channels by category and property group.} Visual Recreation and Footage Editing share trace and semantic-window backends. Channels are compared on the sampled grid fixed by the scoring plan.}
\label{tab:property-dictionary}
\begin{tabularx}{\linewidth}{@{}>{\raggedright\arraybackslash}p{2.1cm}*{4}{>{\raggedright\arraybackslash}X}@{}}
\toprule
\tabhead{Category} & \tabhead{Appearance} & \tabhead{Spatial} & \tabhead{Timing} & \tabhead{Dynamics} \\
\midrule
Visual Recreation, Footage Editing & Content, fill and stroke color, stroke width, opacity & Position, size, crop, relative placement & Correspondence lag, event onsets & Motion direction and speed, motion response and binding \\[3pt]
Motion Fusion & Proportions, material, emissive and shape states, lighting & Centroid, intrinsic and screen size, screen position, contact gap, attachment offset, depth order & Activity and motion onsets, event times & Displacement, speed, angular speed, relative orientation, deformation, attachment drift, surface motion field \\[3pt]
3D Camerawork & Focal length, principal point & Camera position, orientation & Activity events, cut times & Per-frame increments of position, orientation and lens, subject-relative actions \\
\bottomrule
\end{tabularx}
\end{table}

\subsection{Reconstruction Observations}
\paragraph{Object and Source Observations.}
\label{sec:verif-code}
Source and render probes supply observed tracks, properties and events for scoring under the task's plan. Depending on the task, these can include projected position and size, visible opacity, asset identity, time intervals or motion over matched samples. Binding to the actual drawing is checked against the public target definition.

\paragraph{Scene and Camera Observations.}
\label{sec:verif-motion}
Scene probes can provide sampled surface motion, projected geometry or explicit event data. A surface motion field, with object and camera motion separated, supplies Dynamics evidence. Independent Spatial or Timing credit requires a corresponding check in the plan. Camera tasks use executed trajectory and projection records. Scene subjects provide reference geometry unless the task also declares them as scored targets.

\subsection{Structured Comparison}
\label{app:structured-scoring}

\begin{figure}[!t]
\centering
\includegraphics[width=\linewidth]{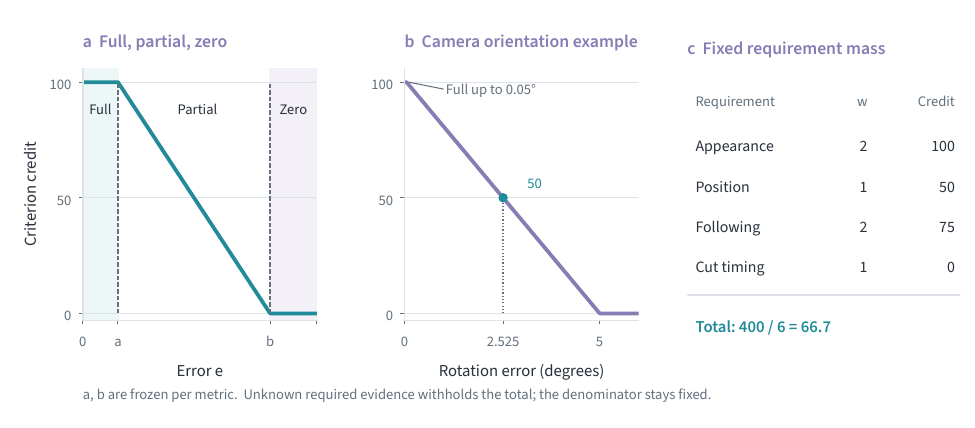}
\caption{\textbf{Credit curves and fixed requirement weights.} (a) A numerical rule gives full credit up to $a$ and zero from $b$, with linear credit between them. (b) The 3D Camerawork orientation check uses $a=0.05^{\circ}$ and $b=5^{\circ}$, yielding a score of 50 at $2.525^{\circ}$. (c) Four illustrative requirement credits aggregate to $(2\cdot100+1\cdot50+2\cdot75+1\cdot0)/6=66.7$. If a required observation remains unresolved, no total score is reported; its weight remains fixed.}
\label{fig:error-to-credit}
\end{figure}

\paragraph{Observation Packets.}
Both packets record the task profile and the digests of the property dictionary and scoring plan. Each channel specifies its unit, coordinate space, clock and sample grid. The records link measured values to execution evidence and document confirmed omissions and unknown values. For relations, they name the endpoints; for events, cameras and timeline requirements, they record the time scope.

Model analysis proposes target definitions and bindings from the reference project and task specification, without access to private reference values. Reviewed definitions are frozen before measurement. A trusted observer confirms which source the run actually executed and measures the supported properties. Records bind the source, observer, runtime, assets and observation grid. The plan fixes admissible correspondences, with one assignment used across the scored properties of a collection. If ambiguous matches can change credit, the affected measurement remains unknown. A permitted member permutation leaves the score invariant.

Missing binding evidence leaves visual presence unresolved until drawing evidence establishes it. A bounding box supports position and extent checks. Glyph identity and contours require their own evidence. A required property that the observer cannot measure in a valid submission remains unknown.

\paragraph{Metrics and Credit.}
Position uses a declared coordinate frame and scale. Rotations compare geodesic angles on each sampled frame, sizes can compare log ratios, and canonicalized categorical values use exact equality. Accepted intervals and credit outside them are specified per criterion.
For a numerical criterion $j$ with error $e\ge0$ in its declared unit, credit is full up to a bound $\tau_j^{\mathrm{full}}$, zero from a tolerance $\tau_j^{\mathrm{zero}}$ and linear in between; most checks set $\tau_j^{\mathrm{full}}=0$, while some category metrics keep a full-credit band:
\begin{equation}
\phi_j(e)=100\begin{cases}
1, & e\le \tau_j^{\mathrm{full}},\\
\dfrac{\tau_j^{\mathrm{zero}}-e}{\tau_j^{\mathrm{zero}}-\tau_j^{\mathrm{full}}}, & \tau_j^{\mathrm{full}}<e<\tau_j^{\mathrm{zero}},\\
0, & e\ge \tau_j^{\mathrm{zero}},
\end{cases}
\label{eq:error-credit}
\end{equation}
and a sampled criterion averages $\phi_j$ over its sample grid. Interval criteria give full credit inside the accepted range and a declared penalty outside it. Ordered event matching uses one-to-one correspondence and accounts for missing and extra events. Each comparator fixes its temporal support and allowed alignment before candidate evaluation.

A parent-relative spatial requirement can penalize an incorrect assembly offset while a dynamics requirement credits correct following. A required contact or cut can receive a dedicated event check. Each distinct obligation counts once in the plan.

\paragraph{Weight Assignment.}
Reference salience accumulates each element's visible area over time, with the per-frame area capped at 5\% of the frame and scaled by opacity; camera shots use their duration. Elements are ranked by salience, and those covering the leading half of cumulative salience are core. Hardness is judged from the reference alone. A camera shot is hard when, within the shot, its vertical field of view changes by at least $7^{\circ}$ while the camera travels at least 1.5\,m, or when its peak angular speed reaches $150^{\circ}$/s. Object requirements are hard when they carry a frozen relation check, or, in 3D scenes, applicable contact, attachment or occlusion. A requirement that is both core and hard receives weight 3.

\paragraph{Score Accounting.}
A relational criterion declares its endpoints as prerequisites: the confirmed absence of an endpoint fails only the dependent criteria, while a wrong relation between present endpoints is a property error. In the notation of \cref{eq:target-accounting}, each criterion accounts for $a_j(100-g_js_j)/\sum_k a_k$ lost points, which locates where a submission fell short. Every requirement declares each property group as scored, not applicable or deferred. Where a plan checks for extra behavior, it applies a bounded deduction to the affected requirement.

\subsection{Creation Assessment}
\label{sec:creation-judging}
The judge is \texttt{gemini-4-argon}. Each request covers one shot and one rubric aspect, such as visual content, text, motion or transitions, and supplies the screenplay, the binary questions of that aspect and continuous clips of the rendered film (\cref{app:script-scoring}). Each question names the objects and the condition it checks. Audio questions are excluded. The verdicts are regrouped into the four property groups, and deterministic code averages the pass rates within each group and then across the four groups (\cref{sec:score-reporting}). Records retain the judge and prompt versions, rubric, decoding settings and render checksum. A failed judge execution leaves the task score unresolved.

\subsection{Human Assessment Protocols}
\label{sec:human-studies}
The two human studies of \cref{sec:future-work} cover agreement with the Creation judge and Expressive Quality.

\paragraph{Creation Calibration.}
The first study tests whether the VLM judge applies screenplay-derived rubrics consistently with human assessment.
Outputs are sampled across Creation tasks, agent systems and observed levels of requirement fulfillment.
Controlled omissions, timing shifts and frozen motion test fault sensitivity, and valid alternative designs test permitted variation.
Calibration examples are kept apart from held-out validation outputs, and the judge prompt, rubric and weights are frozen before validation.

Raters independently inspect the screenplay, rubric and the same evidence supplied to the judge. System identity and automatic decisions are hidden.
Each criterion receives a binary pass/fail judgment, supporting timestamps and a brief rationale.
Inter-rater agreement (Cohen's $\kappa$) is computed before disagreements are adjudicated, and resolved judgments are kept separately.
Against the adjudicated human judgments, the comparison reports criterion-level agreement (Cohen's $\kappa$ and raw agreement), false-pass and false-rejection rates, the Spearman rank correlation of task scores and their mean absolute difference.
Repeated judge evaluations of the same outputs measure judgment stability as the agreement ($\kappa$) between runs.

\paragraph{Expressive Quality.}
The second study assesses whole-film visual expression as a secondary outcome across both tracks.
Sampling covers all five production categories and the represented video purposes.
Raters view the full rendered film with its production purpose and brief, with system identities hidden and presentation order counterbalanced.
Shared rating anchors cover composition, typography, color, visual hierarchy, motion rhythm, stylistic coherence and suitability for purpose.

Human ratings are collected separately from requirement-fulfillment judgments.
We report sampling coverage, eligible output counts, rater agreement and uncertainty for each system and category.
Handling of missing renders is fixed before sampling.
Expressive Quality contributes no weight to the task scores, category means or Overall.
\Cref{tab:human-study-plan} summarizes the study design and measurements to be collected.

\begin{table}[!t]
\centering\small
\caption{\textbf{Human-study reporting plan.} Planned endpoints and result fields for the two human studies.}
\label{tab:human-study-plan}
\resizebox{\linewidth}{!}{%
\begin{tabular}{@{}lll@{}}
\toprule
\tabhead{Study} & \tabhead{Planned reporting} & \tabhead{Measurements to collect} \\
\midrule
Creation calibration & Agreement with humans, false passes and rejections, fault sensitivity, stability & Tasks, outputs, raters, agreement \\
Expressive Quality & System and category ratings, purpose coverage, rater agreement, uncertainty & Outputs, raters, quality scores \\
\bottomrule
\end{tabular}%
}
\end{table}

\FloatBarrier
\section{Task Authoring}
\label{app:construction-details}
\suppressfloats[t]

This appendix describes how the five-stage pipeline of \cref{sec:construction} is applied in each category.
Each category subsection describes how references and grading evidence are constructed, and the last subsection describes quality validation. The composition is given in \cref{sec:overview-scale} and the public specifications in \cref{app:design-details}.

\begingroup\emergencystretch=1.5em
\begin{table}[!t]
\centering\footnotesize
\caption{\textbf{Two reference origins.} Existing videos guide iterative reconstruction, while expert-defined briefs guide implementation and review. Both origins feed the shared authoring pipeline.}
\label{tab:pipelines}
\label{tab:route-steps}
\begin{tabular*}{\linewidth}{@{\extracolsep{\fill}}llll@{}}
\toprule
\tabhead{Origin} & \tabhead{Specification} & \tabhead{Construction} & \tabhead{Review} \\
\midrule
Existing video & \makecell[lt]{Required content, composition,\\motion and timing} & \makecell[lt]{Reconstruct an editable project\\through repeated rendering} & \makecell[lt]{Expert comparison with the film\\and refinement of discrepancies} \\[3pt]
Creative brief & \makecell[lt]{Expert-defined script\\or scene goal} & \makecell[lt]{Agents implement detailed visual,\\camera and motion requirements} & \makecell[lt]{Expert review of successive\\renders and guided revision} \\
\bottomrule
\end{tabular*}
\end{table}

\definecolor{fvCoral}{HTML}{CF6A50}
\definecolor{fvCoralDeep}{HTML}{B5553C}
\definecolor{fvMuted}{HTML}{59666E}

\paragraph{Reference Origins.}
\label{sec:gt-origin}

As \cref{tab:pipelines} shows, both origins produce an editable reference project and a reviewed render from which task-specific observations are obtained, and either origin can support either track.
A reference-solution run and a full-timeline fidelity check then connect the reference to grading (\cref{sec:val-oracle}).
\Cref{fig:strips} shows how reference videos evolve over time in three categories.

\paragraph{Worked Example.}
\label{sec:authoring-example}
For the FV logo case in \cref{sec:assessment-example}, authors specify the required mark, initial position, colors and 2-to-3-second interval, then export the reference's relevant states and change profiles.
Controls omit the logo, change its colors, shift its onset or freeze its movement, and valid alternatives test whether the pass conditions preserve the brief's design freedom.

\begin{figure}[!t]
\centering
\newcommand{\strip}[3][]{\foreach \i in {0,...,4}{\shot[#1]{0.19\linewidth}{strips/#2-\i.jpg}\ifnum\i<4\hfill\fi}\\[1pt]{\footnotesize #3}\\[5pt]}
\includegraphics[width=0.19\linewidth]{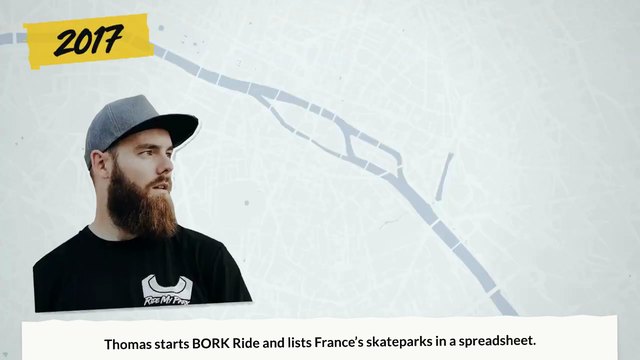}\hfill\includegraphics[width=0.19\linewidth]{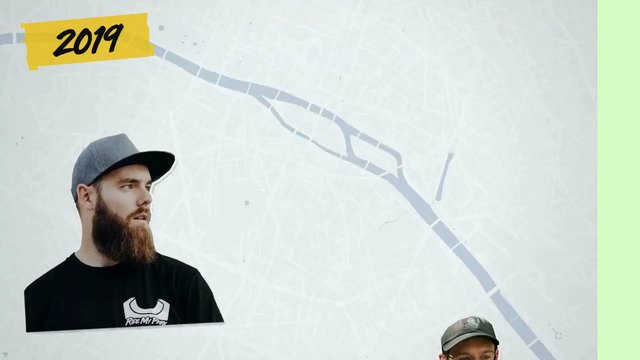}\hfill\includegraphics[width=0.19\linewidth]{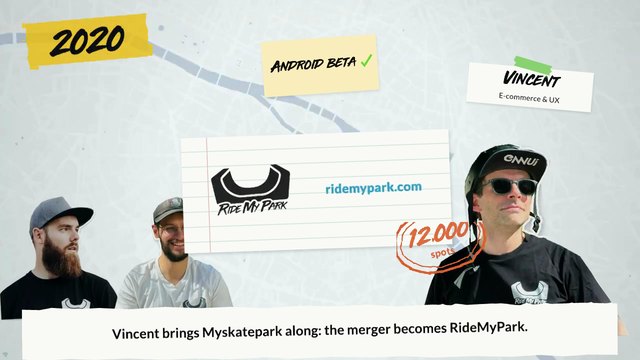}\hfill\includegraphics[width=0.19\linewidth]{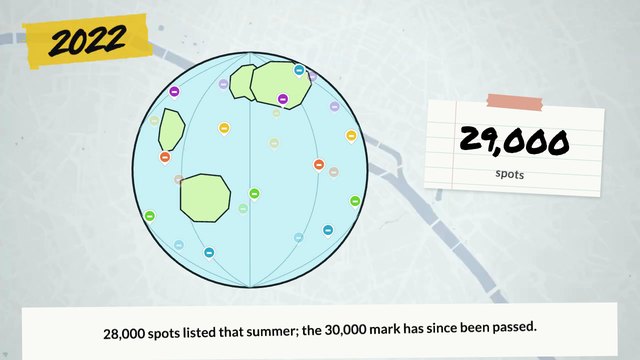}\hfill\includegraphics[width=0.19\linewidth]{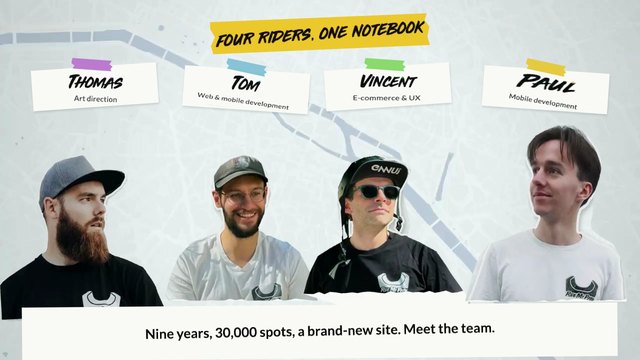}\\[1pt]{\footnotesize (a) \catfootage, \texttt{reconstruct-footage-founders}: 1280$\times$720, 51.6\,s; frames at 5.7, 15.5, 25.8, 35.1 and 45.5\,s}\\[5pt]
\includegraphics[width=0.19\linewidth]{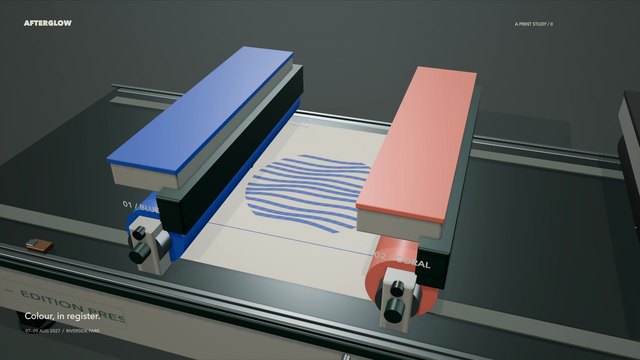}\hfill\includegraphics[width=0.19\linewidth]{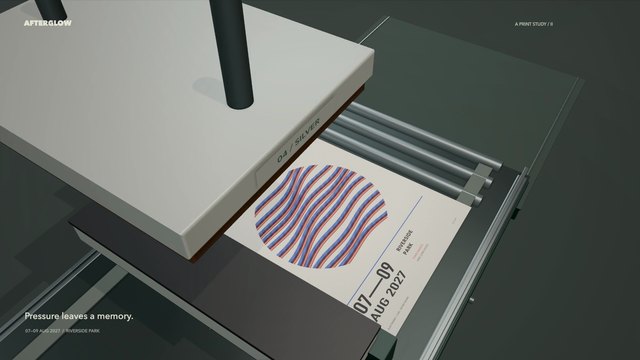}\hfill\includegraphics[width=0.19\linewidth]{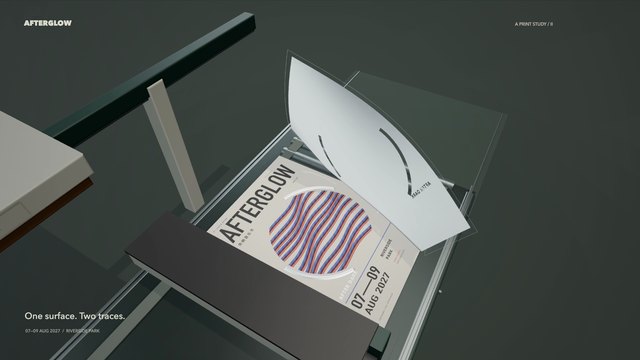}\hfill\includegraphics[width=0.19\linewidth]{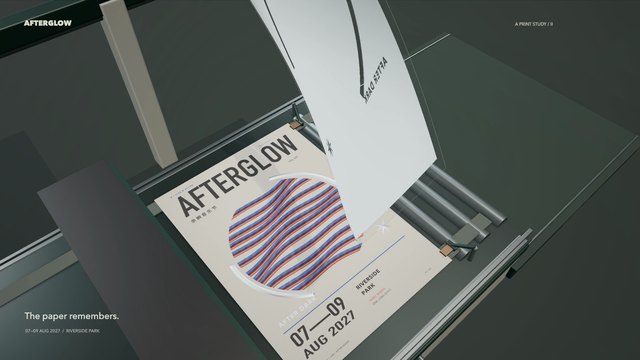}\hfill\includegraphics[width=0.19\linewidth]{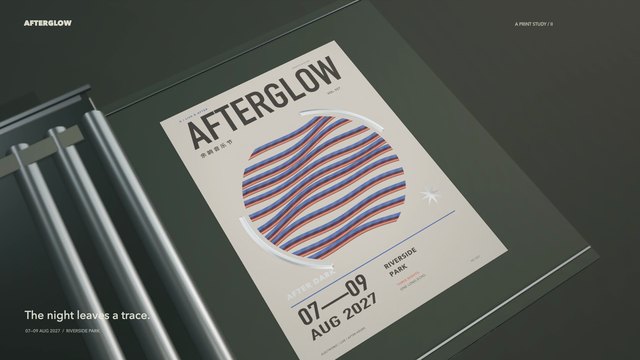}\\[1pt]{\footnotesize (b) \catmotion scene description, \texttt{reconstruct-motion-afterglow-foil-peel}: 1920$\times$1080, 10\,s, rendered from the GT; frames at 1.3, 3.0, 5.5, 6.5 and 9.5\,s}\\[5pt]
{\def\fvshotratio{0.28125}\includegraphics[width=\linewidth]{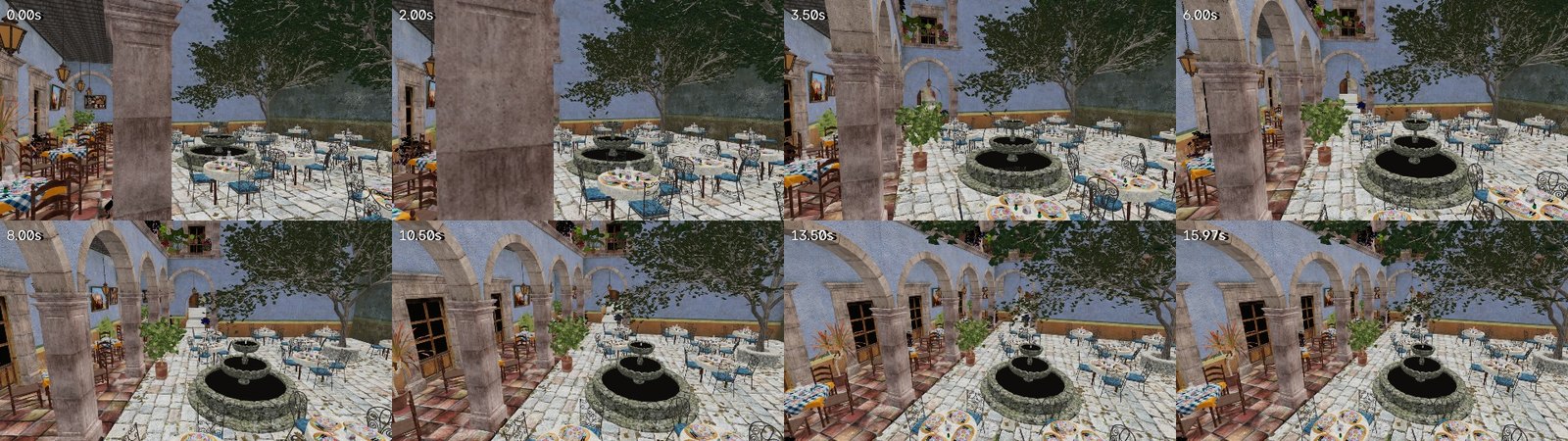}}\\[1pt]
{\footnotesize (c) \catcamera, \texttt{reconstruct-camera3d-san-miguel-long-take}: storyboard of the accepted 1280$\times$720, 16\,s director reference, rendered from the GT camera}
\caption{\textbf{Reference videos over time.} Footage tasks change shots and layouts, motion tasks evolve one dense scene, and camera tasks move through a fixed 3D scene. Frames (a) and (b) are sampled from the reference videos in the task packages; (c) is taken from the accepted director reference.}
\label{fig:strips}
\end{figure}

\subsection{Visual Recreation}
\label{sec:pipe-visual-source}
\noindent\textbf{Reference Construction.} References are motion-graphics films with editable Remotion source, drawn from public repositories, owner-provided projects and local remakes. Imported projects are adapted to the pinned runtime through local font shims and starter entry points.

\noindent\textbf{Grading Evidence.} Grading uses observed element traces or source annotations with private checklists. Tasks with source-property checks also expose object tables and, where supplied, masked anchor frames such as those in \cref{fig:anchors}. Dense compositions require the agent to recover many coordinated entrances, layouts and motion channels (\cref{sec:verif-code}).

\subsection{Footage Editing}
\label{sec:pipe-footage}
\noindent\textbf{Reference Construction.} Tasks are built around supplied photographs, illustrations, textures, fonts and other independent media. A human author chooses the excerpt, asset boundaries and acceptance target, and the reference follows either authored choreography or iterative reconstruction of an existing film. Motion is attached to objects and parent groups through interpolation, easing, springs, paths and transitions, while full-frame scene plates, reference playback and per-frame tracing are prohibited. Scripts retrieve the media, render the reference solution and export observations.

\noindent\textbf{Grading Evidence.} The public asset inventory identifies files and checksums, and agents identify the permitted assets and implement the composition. The challenge is to recover repeated instances, cropping, layering, group motion and event timing from the reference.

\subsection{Motion Fusion}
\label{sec:pipe-gt-films}
\label{sec:pipe-scene-json}
\noindent\textbf{Authored Scenes.} A human expert sets the design scope, agents propose and implement the scenes in Remotion with three.js scenes and DOM text, and the expert reviews the completed films. The frozen reference package contains a deterministic project, sampled object states and a measurement sheet exported from the design and code. Agents review keyframes for up to two rounds before the sheet is self-tested and hash-frozen. Public object cards identify the targets, and submitted three.js objects carry their card IDs in \texttt{userData}, while designs, event answers and measurement sheets stay private. The distinctive challenge is coordinating motion, attachments, reflections and visibility within one scene.

\noindent\textbf{Scene Descriptions.} Procedural three.js scenes probe continuous deformation, contact-driven deformation and contention for shared resources. The submission is a \texttt{scene.json} containing geometry, materials, camera and per-frame state, built against a supplied schema, example and trusted three.js viewer. A patch exposes the GT scene, camera and renderer to a probe that samples surface points on every frame at verifier runtime, and the verifier compares sampled visible motion with camera and object contributions separated (\cref{sec:verif-motion}).

\subsection{3D Camerawork}
\label{sec:pipe-camera-hard}
\label{sec:pipe-camera-templates}
\noindent\textbf{Camera Recovery.} Each task supplies a target film and a fixed third-party scene, such as San Miguel from \href{https://casual-effects.com/data/}{Morgan McGuire's Computer Graphics Archive} or UNC Power Plant. The authored camera is recorded during rendering; designed cases can use per-shot \href{https://docs.scipy.org/doc/scipy/reference/generated/scipy.interpolate.PchipInterpolator.html}{shape-preserving cubic interpolation (PCHIP)}. Task-local generators reproduce the camera and export retimed conditions, depth grids and applicable subject samples for private grading. Public tools include scene data, licenses, reference frames, a proxy renderer and a static gate checker, and abrupt shot changes, changing field of view and moving subjects make recovery demanding.

\noindent\textbf{Screenplay Templates.} Multi-shot films are designed from the eight video purposes in scenes drawn from Blender demos, NVIDIA ORCA and Sketchfab. Agents encode subject-motion and camera-motion rules, scripts check physics and clearance, and the owner reviews trial renders. The fixed rules generate cameras, cuts, subject states and depth caches for base, retimed and applicable subject variants, and a separately authored reference solution recovers the generator's structure. Agents receive the named moving objects and a baked approximation of the path-traced scene, with geometry, subject motion and projection shared. The demands include following subjects with lag, shifting focus between subjects and coordinating field-of-view changes with camera motion.

\subsection{Script Realization}
\label{sec:pipe-creation}
Authors translate the screenplay into binary rubric items as described in \cref{sec:construction}, separating required subjects, actions, relations, shot order, transitions and timing from permitted choices of assets, composition and motion. A feasible reference implementation shows that the brief can be realized without prescribing a unique visual answer. The judging protocol is described in \cref{sec:creation-judging}.

\paragraph{Creation Example.}
\emph{Cat \& Coffee}, one of the 20 Script Realization tasks, follows a cat and a drinker through 13 art styles while keeping the scene composition consistent. The closing shot uses flat vector illustration and ends with the drinker petting the cat. \Cref{tab:creation-rubric-example} illustrates how visual and temporal requirements in the public brief become binary checks.
\begin{table}[!t]
\centering\footnotesize
\caption{\textbf{Example rubric checks for Cat \& Coffee.} These illustrative binary checks use continuous clips of the closing shot.}
\label{tab:creation-rubric-example}
\begin{tabularx}{\linewidth}{@{}p{0.42\linewidth}X@{}}
\toprule
\tabhead{Screenplay obligation} & \tabhead{Binary criterion and evidence} \\
\midrule
The year and style subtitle appear in the upper right. & Pass if the closing clip shows \emph{2026} above \emph{Flat Vector Illustration} in the upper right. \\
The cat jumps onto the left side of the table around 12\,s. & Pass if the clip near 12\,s shows the cat jump and land on the left side of the tabletop. \\
The drinker pets the cat from 13.4 to 15.0\,s. & Pass if the clip shows the drinker's hand petting the cat during that interval. \\
\bottomrule
\end{tabularx}
\end{table}

\subsection{Quality Validation}
\label{app:validation-details}
\label{sec:val-oracle}
\label{sec:validation-controls}
Quality validation applies stage (e) of \cref{sec:construction} to every task. A reference-solution run checks agreement with the grading target; when the solution also supplies the GT, this is a consistency check. A fidelity check confirms the re-render preserves the intended video and the scored requirements, and an independently authored submission under the public contract tests whether the evaluator accepts valid implementation choices.

The structured comparator is calibrated in three steps.
\begin{enumerate}
\item Check the reference render and valid alternatives, such as equivalent transforms, member permutations and alternative geometry that preserves the declared targets and public bindings.
\Needspace{4\baselineskip}
\item Introduce graded faults in one requirement at a time. Use blinded review of renders to identify accepted deviations and clear failures, which inform the full-credit and zero-credit bounds. If the measured errors for those groups overlap, revisit the metric or its observation scope.
\Needspace{3\baselineskip}
\item Freeze the observer, support, bounds and weights. Use held-out tasks to test fault sensitivity, retention of credit for valid alternatives, repeatability and unresolved observation rates.
\end{enumerate}
Empty, unchanged-starter and frozen controls probe missing content and motion where applicable. Contact, cuts and reveals can use dedicated event or interval checks when averaging would hide an essential moment. Calibration of the Creation judge is part of the human studies in \cref{sec:human-studies}.

\FloatBarrier
\section{Experimental Details}
\label{app:trial-details}

\subsection{Task Instructions}
One generator renders every task instruction from three layers of constraints: general rules shared by all tasks, the category's goal and technical rules, and optional task-specific notes. Each instruction has six sections, and it states the public contract without describing how submissions are scored. The abridged Visual Recreation instruction below shows the structure; braces mark task-specific values.
\begin{promptbox}[Task instruction, Visual Recreation (abridged)]
\small
\textbf{Task.} Recreate the reference video as an editable Remotion project, using the provided assets and starter project. Output specification: \{W\}$\times$\{H\}, \{fps\} fps, \{N\} frames.\par
\textbf{Inputs.} \texttt{materials.json} lists every provided file with its role, checksum and permitted use; the reference video, starter project, input guide, asset list and object table follow.\par
\textbf{Constraints.} \emph{General:} use only the provided inputs and do not access private files or the network; the reference video and frames are for observation only; the output must be generated deterministically by the submitted project. \emph{Visual Recreation:} recreate the reference's overall visual appearance, structure and rhythm through code; build with the provided Remotion project and locked libraries and do not pre-render imagery with other tools; use only the provided assets as raster media and write text, shapes and vector graphics as editable code; express motion as functions of time rather than per-frame values; put a unique \texttt{data-scene-object} on the node that is actually rendered. \emph{This task:} \{task-specific notes\}.\par
\textbf{Deliverables.} The complete editable project with \texttt{PROJECT.md} documenting its entry, and \texttt{output.mp4} rendered from it.\par
\textbf{Tools and Runtime.} \texttt{task-render} for the full export; 4 CPU cores, 8\,GiB memory and a six-hour limit including rendering, with no network access.
\end{promptbox}

\subsection{Script Realization Scoring}
\label{app:script-scoring}
The Script Realization scores in \cref{tab:controlled-results,fig:leaderboard-layout-preview} come from 20 screenplay tasks. Each system ran each task once. Claude Opus~5.5, GPT-6.1 Sol, GLM~5.3 and DeepSeek~V4.1 Flash come from the original study, which ran six systems (120 runs); MiMo V2.6 Pro, Kimi K3 and Step 5 Preview were later run on the same tasks and judged with the same questions. The judge, \texttt{gemini-4-argon}, answered 1,867 binary questions per system. Each request covered one shot and one aspect, with continuous clips of at most 12\,s that overlap by 0.5\,s, sampled at 24\,fps at original speed, and an output limit of 4,096 tokens. The binary verdicts were then regrouped into the four property groups without judging the films again. The original rubrics of ten tasks also contain 234 audio questions; these are excluded, so audio does not enter the scores.

\subsection{Earlier Protocol Controls}
\label{app:earlier-trials}
Two controls that predate the controlled comparison show how a score overstates fidelity when its measurements miss part of the film. For the \texttt{diffusion} task, a score of 100 on source-property checks coexists with a frame-level score of 56.5, below the 61.5 of the unmodified starter project. For the \texttt{claim} task, a trace with stationary objects and the original camera motion scores 99.37 under its earlier protocol. These results apply to the task versions and metrics reported and, like the other controls, remain outside the benchmark aggregate.

\FloatBarrier
\Needspace{20\baselineskip}
\section{Reproducibility}
\label{app:repro}

\paragraph{Frozen inputs and runtime.}
Record task and scorer versions, input and code hashes, image and dependency locks, hardware, render settings and agent configuration. For Reconstruction, bind the schema, property dictionary, observer and scoring plan and observe both projects in the same pinned verifier environment. To reproduce a reported score, use the scorer and plan bound to its archived run. Agent and verifier budgets are six hours each. Both run without internet access, except that the agent reaches its model endpoint.

\paragraph{Version identifiers.}
The main text and appendices refer to scoring rules, task versions and studies by description. \Cref{tab:version-map} maps each description to the identifier used in the repository and run records.

\begin{table}[!b]
\centering\footnotesize
\caption{\textbf{Version identifiers.} Descriptions used in the text and the identifiers recorded in the repository and run archives.}
\label{tab:version-map}
\begin{tabularx}{\linewidth}{@{}>{\raggedright\arraybackslash}X>{\raggedright\arraybackslash}p{3.1cm}>{\raggedright\arraybackslash}X@{}}
\toprule
\tabhead{Description in text} & \tabhead{Where} & \tabhead{Identifier} \\
\midrule
Reconstruction scoring protocol & \cref{sec:score-reporting} & \texttt{reconstruction-four-attributes-v2} \\
3D Camerawork orientation check & \cref{fig:error-to-credit}b & \texttt{camera-quality-v3}, with tolerances from \texttt{camera-quality-v2} \\
Script Realization study & \cref{tab:controlled-results,app:script-scoring} & Bench20 delivery of 2026-11-05 \\
Public object evidence & \cref{fig:anchors} & \texttt{reconstruct-visual-autumn-post}, task input 0.5.1 \\
Reference videos over time & \cref{fig:strips}a, c & \texttt{reconstruct-footage-founders}, recorded task set 0.1.6; San Miguel director reference of 2026-10-16 \\
DeepSeek V4.1 Flash & \cref{sec:experiment-setup} & model release listed as DeepSeek V4.1 Flash 1017 \\
\bottomrule
\end{tabularx}
\end{table}

\paragraph{Task execution.}
Restore registered assets and hidden data by checksum. The corresponding runs require private input access. Use the pinned checkout to restore the task and run the controls.
\begin{promptbox}[Task setup and controls]
\begin{verbatim}
uv sync
uv run python scripts/media/media.py setup --task <task-id>
uv run python scripts/media/media.py check --task <task-id>
uv run harbor run -p tasks/<task-id> -a oracle -e docker -o jobs
uv run harbor run -p tasks/<task-id> -a nop -e docker -o jobs
\end{verbatim}
\end{promptbox}
Run the agent with its recorded settings and retain the source, outputs, logs and receipts.

\paragraph{Structured comparison.}
The \href{https://github.com/GeminiLight/FrontierVideo/blob/76958c9c8cfb638011fdaa7313f266f48d940c13/skills/frontiervideo-json-extraction/SKILL.md}{JSON export guide} recovers the paired observation packets of the submission and the reference from verified runs, with \href{https://github.com/GeminiLight/FrontierVideo/blob/76958c9c8cfb638011fdaa7313f266f48d940c13/skills/frontiervideo-json-extraction/references/capture-routes.md}{per-category instructions} for projects whose capture is incomplete. Retain both packets, object correspondences, the fixed scoring plan and comparison record. Creation keeps its evidence and judge records (\cref{sec:creation-judging}). See the \href{https://github.com/GeminiLight/FrontierVideo/blob/1384651f51600e2c39ca56c913ae95f0121e414f/research/paper/sections/E-reproduction.tex}{historical reproduction records} for earlier settings.

\end{document}